\documentclass[lettersize,journal]{IEEEtran}
\usepackage{amsmath,amsfonts}
\usepackage{algorithmic}
\usepackage{algorithm}
\usepackage{array}
\usepackage[caption=false,font=normalsize,labelfont=small,textfont=small]{subfig}
\usepackage{textcomp}
\usepackage{stfloats}
\usepackage{url}
\usepackage{verbatim}
\usepackage{graphicx}
\usepackage{cite}
\usepackage{multirow}
\usepackage{pifont}
\usepackage{color}
\newcommand{\edita}[1]{\textcolor{black}{#1}}
\newcommand{\editb}[1]{\textcolor{black}{#1}}

\begin{document}

\title{SRC-Net: Bi-Temporal Spatial Relationship Concerned Network for Change Detection}

\author{Hongjia Chen,
Xin~Xu,
Fangling Pu,~\IEEEmembership{Member,~IEEE}
\thanks{© 2024 IEEE.  Personal use of this material is permitted.  Permission from IEEE must be obtained for all other uses, in any current or future media, including reprinting/republishing this material for advertising or promotional purposes, creating new collective works, for resale or redistribution to servers or lists, or reuse of any copyrighted component of this work in other works. (Digital Object Identifier 10.1109/JSTARS.2024.3411622)}
\thanks{This work was supported in part by National Natural Science Foundation of China under Grant 62071336 and in part by National Natural Science Foundation of China under Grant 62271356. (\textit{Corresponding author: Xin Xu)}}
\thanks{The authors are with the Collaborative Sensing Laboratory, Electronic Information School, Wuhan University, Wuhan 430079, China (e-mail: chj1997@whu.edu.cn; xinxu@whu.edu.cn; flpu@whu.edu.cn).}}

\maketitle

\begin{abstract}
Change detection (CD) in remote sensing imagery is a crucial task with applications in environmental monitoring, urban development, and disaster management.
\edita{CD involves utilizing bi-temporal images to identify changes over time.
The bi-temporal spatial relationships between features at the same location at different times play a key role in this process.
However, existing change detection networks often do not fully leverage these spatial relationships during bi-temporal feature extraction and fusion.
In this work, we propose SRC-Net: a bi-temporal spatial relationship concerned network for CD.}
\edita{The proposed SRC-Net includes a Perception and Interaction Module that incorporates spatial relationships and establishes a cross-branch perception mechanism to enhance the precision and robustness of feature extraction.
Additionally, a Patch-Mode joint Feature Fusion Module is introduced to address information loss in current methods. It considers different change modes and concerns about spatial relationships, resulting in more expressive fusion features.}
\edita{Furthermore, we construct a novel network using these two relationship concerned modules and conducted experiments on the LEVIR-CD and WHU Building datasets.
The experimental results demonstrate that our network outperforms state-of-the-art (SOTA) methods while maintaining a modest parameter count.}
We believe our approach sets a new paradigm for change detection and will inspire further advancements in the field.
The code and models are publicly available at \url{https://github.com/Chnja/SRCNet}.
\end{abstract}

\begin{IEEEkeywords}
Change detection (CD), deep learning, optical remote sensing images, spatial relationship
\end{IEEEkeywords}

\section{Introduction}
\IEEEPARstart{C}{hange} detection (CD) aims to identify change areas in the study region through analysis of multiple remote sensing observations \cite{singh1989review}, and has gradually become a research hotspot in remote sensing \cite{li2020amn,yang2021da2net}.
By comparing remote sensing images taken at different times of the same region, change areas can be extracted on a pixel-by-pixel map.
CD has played a significant role in many fields including environment monitoring \cite{reba2020systematic,sippel2020climate,sun2022understanding}, resource management \cite{sim2006grid}, urban planning \cite{ji2018fully}, and disaster assessment \cite{lei2019landslide,peng2020optical}.

In recent years, with the widespread adoption of deep learning, particularly the advancements in Convolutional Neural Networks (CNNs) \cite{lecun1998gradient} and Transformers \cite{vaswani2017attention}, CD has gradually transitioned from traditional Pixel-based and Object-based methods \cite{hussain2013change,lambin1994change,celik2009unsupervised,marchesi2009ica} to two-stage deep learning approaches, and finally to the end-to-end deep learning networks \cite{saha2020unsupervised,saha2020building}.
However, most existing CD networks have originated from image semantic segmentation, 
\editb{lacking structure design for perceiving bi-temporal spatial relationships.
Therefore, existing methods are insufficient in the utilization of bi-temporal spatial relationships, which is crucial for CD.}

\editb{In this paper, we define the corresponding relationships that exists between features at the same location at different times as bi-temporal spatial relationships.}
The optical characteristics obtained from imaging the same location at different times should exhibit corresponding relationships.
Similarly, the depth features extracted from the aforementioned optical characteristics should also exhibit corresponding relationships.
Current network structures are rarely designed to take advantage of bi-temporal spatial relationships.
This is similar to using fully connected neural networks (FCs) to analyze images before CNNs were proposed.
Fortunately, networks could perceive partial bi-temporal spatial relationships with extensive learning.
This is similar to how FCs can also perceive some spatial information in images, despite the difficulty involved.

\begin{figure}[t]
\centering
\subfloat[]{
    \includegraphics[width= 0.21\linewidth]{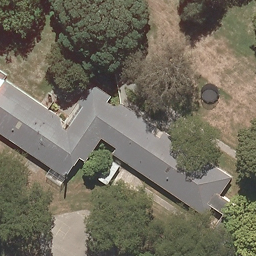}
}
\subfloat[]{
    \includegraphics[width= 0.21\linewidth]{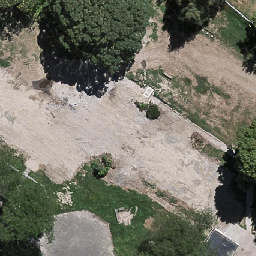}
}
\subfloat[]{
    \includegraphics[width= 0.21\linewidth]{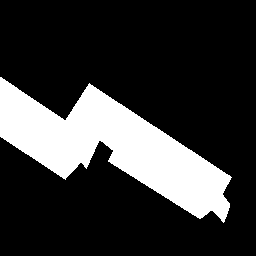}
}
\subfloat[]{
    \includegraphics[width= 0.21\linewidth]{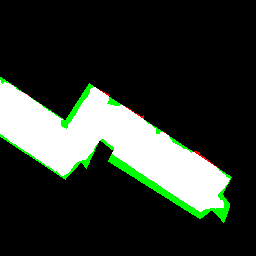}
}

\subfloat[]{
    \includegraphics[width= 0.21\linewidth]{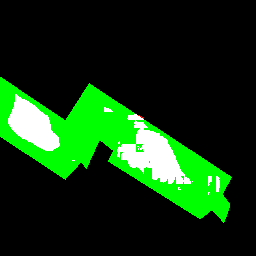}
}
\subfloat[]{
    \includegraphics[width= 0.21\linewidth]{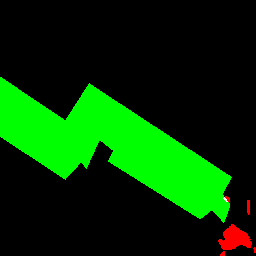}
}
\subfloat[]{
    \includegraphics[width= 0.21\linewidth]{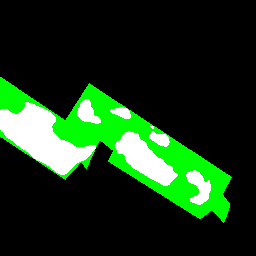}
}
\subfloat[]{
    \includegraphics[width= 0.21\linewidth]{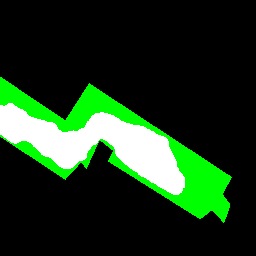}
}
\caption{Bi-temporal remote sensing images.
(a) (b) Original images. (c) Ground truth.
(d) The result of our SRC-Net.
(e) The result of the single-branch network (RDPNet).
(f) The result of the dual-branch network without cross-branch interaction (FC-Siam-diff).
(g) (h) The results of dual-branch network with cross-branch interaction (BIT, DSIFN).
The false positives and false negatives are indicated in red and green, respectively.
Other colors represent true positives.}
\label{obser}
\end{figure}

Existing network structures can be categorized into two groups: single-branch networks and dual-branch networks.
In single-branch networks, the bi-temporal inputs are concatenated into one matrix as distinct channels, which enables semantic segmentation networks to address the CD task.
In dual-branch networks, the bi-temporal inputs are processed through a siamese network with shared network parameters to extract distinct feature maps for each temporal, subsequently fusing these feature maps for the change area prediction.
Intuitively, CD should analyze the ground object features at each time, and then compare the corresponding features between different times to detect changes.
\editb{This is similar to the process of dual-branch networks that first extract features from the bi-temporal inputs separately and then compare the differences between the bi-temporal feature maps.}
Besides, the connection operations within a single-branch network lead to the disappearance of correspondence relationships among bi-temporal features, thereby losing some bi-temporal spatial relationships.
This inevitably gives the impression that dual-branch networks should have advantage over single-branch networks.
However, \editb{in the early-used dual-branch networks \cite{daudt2018fully}, each branch extracts features independently, and there is no cross-branch interaction.}
These networks did not perform well in terms of actual results and were significantly inferior to the single-branch networks, as shown in Fig. \ref{obser}(e) and (f).

\begin{figure}[htb]
\centering
\includegraphics[width= 0.75\linewidth]{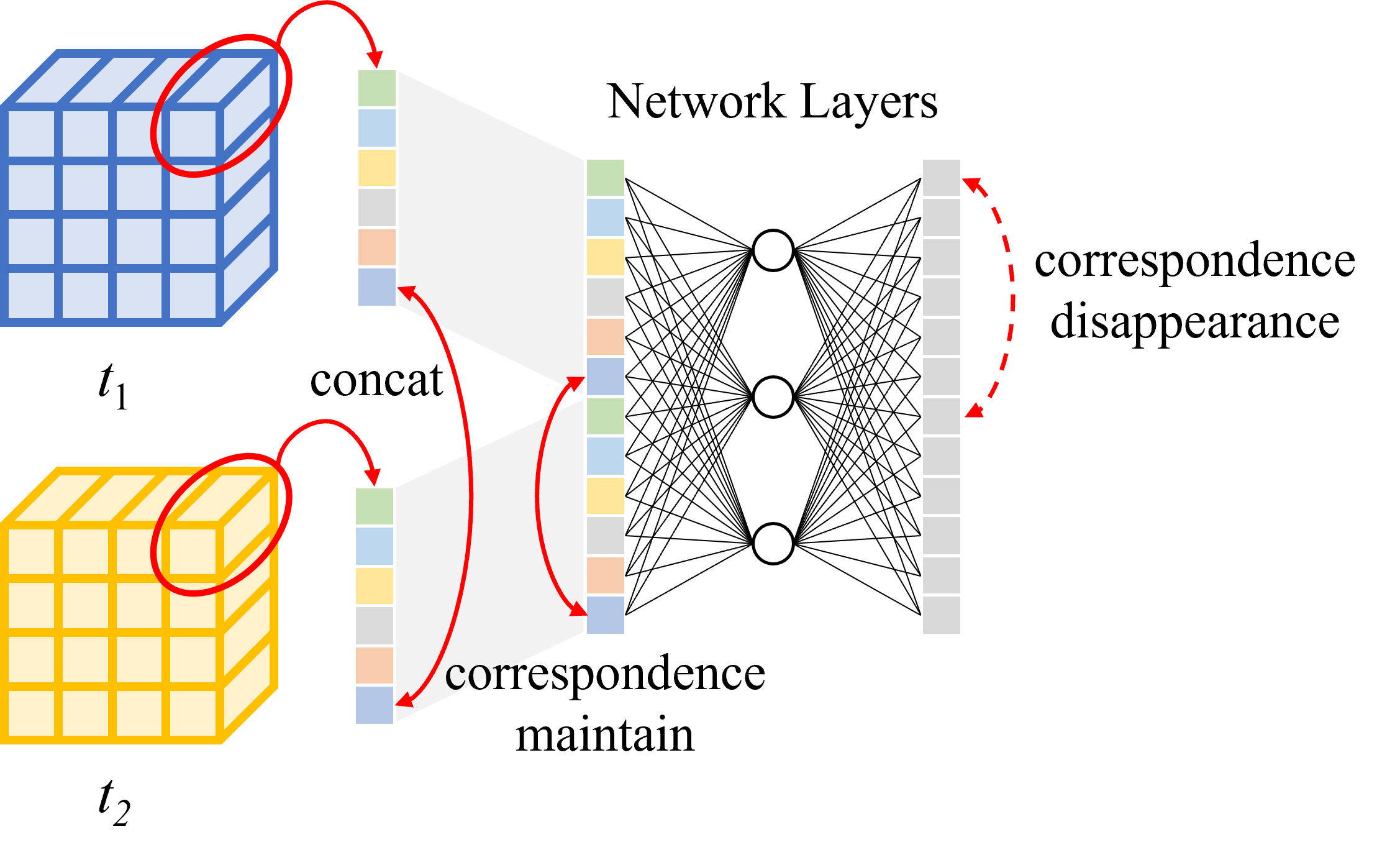}
\caption{\edita{Correspondence disappearance in feature concatenation.}}
\label{single-branch}
\end{figure}

\edita{We argue that this is due to the initial dual-branch networks' inadequate utilization of the bi-temporal spatial relationships during the processes of bi-temporal feature extraction and fusion.
During the feature extraction, the initial dual-branch networks independently extract features in each branch, which lack cross-branch interaction.
Single-branch networks can achieve better results because they concatenate bi-temporal inputs at the beginning. The network does not have the concept of bi-temporal phases and can perceive information across them.
Accordingly, researchers have designed several modules for cross-branch feature interaction.
However, most of them concatenate bi-temporal features and acquire features through network layers.
\editb{As shown in Fig. \ref{single-branch}, there should be correlation between features at the same location at different times, and after concatenating them, the next network layers have no spatial awareness in the dimension of depth, and will treat bi-temporal features as a new feature vector, and ignore the relationship between the different temporal features.}
Therefore, the correspondence relationships among the bi-temporal features would be disrupted.
As shown in Fig. \ref{obser}(g) and (h), while there has been some improvement, there remains substantial room for progression.
In the feature fusion stage, existing approaches can be broadly categorized as module-based methods and subtraction-based methods.
Existing module-based methods have the concatenation mentioned in Fig. \ref{single-branch}, resulting in the corresponding disappearance.
While subtraction-based methods would lead to the features of unchanged regions approaching zero, causing a significant loss of information.
Moreover, current methods have not accounted for the \editb{variability of different changes}.
These turn out to be bottlenecks of CD.}

\edita{In order to tackle the previously mentioned concern, we propose a cross-branch interaction module within the feature extraction phase, along with a bi-temporal feature fusion module.
Firstly, we propose the Perception and Interaction module, which constructs a dual-branch communication mechanism that can perform cross-branch perception without losing bi-temporal spatial relationships, thereby extracting more precise and robust features.
Secondly, we propose a Patch-Mode joint fusion module.
With this module, we prevent the correspondence disappearance in recent fusion modules and the information loss in subtracting methods.
We slice each patch in the bi-temporal feature data into smaller mini-patches, and generate different fusion results for different change modes.
By considering different changing modes and bi-temporal spatial relationships, we could obtain more expressive fusion features.
Finally, we have further optimized the network's backbone and redesigned its architecture based on our previous research RDP-Net\cite{chen2022rdpnet}.
We believe that our work offers a novel approach to utilizing the bi-temporal spatial relationships in multi-temporal remote sensing data, thereby enhancing the information mining capability of such data.}

The SRC-Net has achieved a better performance in the field of remote sensing CD. The major contributions of SRC-Net can be summarized as follows.

\begin{enumerate}
\item{
\edita{We propose a Perception and Interaction Module that utilizes the bi-temporal spatial relationships among bi-temporal inputs. It establishes a cross-branch interaction mechanism during the feature extraction process, leading to the extraction of more precise and robust features.}}
\item{
\edita{We propose a Patch-Mode joint Feature Fusion Module that prevents the information loss present in current methods. Meanwhile, it considers different change modes and concerns about bi-temporal spatial relationships, thereby obtaining more expressive fusion features.}}
\item{
\edita{Our proposed SRC-Net has been evaluated on two public datasets, LEVIR-CD and WHU Building datasets.
The experimental results indicate that the proposed modules utilize bi-temporal spatial relationships effectively, and our network achieved state-of-the-art (SOTA) performance.}}
\end{enumerate}

This article is organized as follows.
\edita{Section II provides a review of the pertinent literature concerning CD networks, feature interaction, and feature fusion.}
Section III describes the CD method proposed in this article. Section IV presents a set of quantitative comparisons and analyses based on experimental results. Finally, the conclusion is presented in Section V.

\section{Related Work}

\subsection{Change Detection Networks}

Neural networks have recently demonstrated promising outcomes in various geoscience applications \cite{tewkesbury2015critical}, including scene classification \cite{yang2021da2net}, water quality estimates \cite{hu2023a2dwqpe}, etc.
NN has also displayed impressive capabilities in CD \cite{chen2019change,wu2021unsupervised,zhu2017change}.
Daudt \textit{et al.} utilized U-Net as a foundation and introduced Early Fusion (EF) and Siamese (Siam) \cite{daudt2018urban}. They designed and proposed FC-EF, FC-Siam-conc, and FC-Siam-diff \cite{daudt2018fully}, which are widely acknowledged as the cornerstone of deep learning-based change detection.
Peng \textit{et al.} \cite{peng2019end} and Fang \textit{et al.} \cite{fang2021snunet} proposed UNet++\_MSOF and SNUNet-CD based on FC-Siam-conc, replacing U-Net with U-Net++ \cite{zhou2018unet++} and merging multiple side outputs of U-Net++.
Papadomanolaki \textit{et al.} \cite{9352207} presented an architecture similar to U-Net, referred to as L-UNet, which incorporates fully convolutional LSTM blocks at each encoding level to capture the temporal relationship of spatial feature representations.
Chen \textit{et al.} \cite{chen2020spatial} introduced a spatial-temporal attention mechanism to improve the discriminative capacity of deep features extracted from bi-temporal images and proposed a late-fusion method STANet.
As another attention-based method, Chen \textit{et al.} \cite{9491802} proposed a bitemporal image transformer (BIT) to accurately and efficiently capture contextual information within the spatial-temporal domain.
Zhang \textit{et al.} \cite{zhang2020deeply} introduced DSIFN, where they designed a shared deep feature extraction network (DFEN) as the encoder and a difference discrimination network (DDN) as the decoder, demonstrating impressive performance.
Chen \textit{et al.} \cite{chen2022rdpnet} proposed a region detail preserving network RDP-Net, avoiding information loss and improving the network’s attention to details.
\editb{Aiming at the loss of the spatial features of buildings caused by the multiple successive down-sampling operations and the problem of incomplete buildings and the blurred edges caused by the complex scenes, Wang \textit{et al.} proposed HDANet \cite{wang2022high} and W-Net \cite{wang2023double}.}
\editb{Zhou \textit{et al.} \cite{10123995} propose a context aggregation network (CANet) to mine interimage context over all training images for further enhancing intraimage context.
Hang \textit{et al.} \cite{10454002} propose an ambiguity-aware network (AANet) to address the difficulties in dealing with ambiguous regions, where pseudo-changes happen or real changes are corrupted.}
\edita{Existing networks have often overlooked the bi-temporal spatial relationships among bi-temporal remote sensing data, which is the central motivation of our work.}

\subsection{Feature Interaction}

\edita{There are two main ideas for dual-branch feature interaction.
One utilizes a module to acquire dual-branch outputs from dual-branch inputs through neural layers, actualizing interaction among the branches via structure design.
The other directly exchanges along spatial or feature dimensions of dual-branch feature maps to actualize dual-branch interaction.}

\edita{The first idea has found considerable application in domains such as multimodal correlation tasks \cite{nguyen2018improved,chen2020bi}, registration \cite{wu2021feature}, matching \cite{wei2020multi}, etc.
Lu \textit{et al.} \cite{lu2016hierarchical} introduced co-attention to aggregate different features from different branches and then distribute them as attention maps respectively.
Yu \textit{et al.} \cite{yu2020deformable} proposed SiamAttn to compute deformable self-attention and cross-attention from dual branches for object tracking.
Wu \textit{et al.} \cite{wu2021feature} proposed FIRE-Net, which aims to investigate the interaction of features within source and target point clouds across various levels.
Chen \textit{et al.} \cite{chen2022mixformer} proposed MixFormer to achieve efficient feature interaction among windows and dimensions.
Some recent networks, such as STANet, BIT and BAN \cite{li2024new}, already employ similar modules.
These methods derive output results through computations across multiple network layers, which often require a considerable amount of computation.
Meanwhile, they would result in the disappearance of the correspondence between bi-temporal features, as shown in Fig. \ref{single-branch}.}

\edita{The other idea is directly exchanging along spatial or feature dimensions of dual-branch feature maps.
CEN proposed by Wang \textit{et al.} \cite{wang2020deep} is a representative, parameter-free multimodal fusion framework. This framework facilitates the dynamic exchange of channels among sub-networks of different modalities. The process of channel exchange is self-guided by the importance of individual channels.
Fang \textit{et al.} \cite{fang2023changer} proposed MetaChanger, directly exchanging features alternately in the spatial and channel dimensions.
Such methods directly exchange dual-branch feature maps along spatial or feature dimensions, which would break the intrinsic relationships within the feature maps, and have not yet achieved widespread adoption.}

\edita{In this paper, we try to combine these two ideas, reducing the computation to an acceptable size without breaking the relationships within the original feature maps.
Therefore, we propose a perception and interaction module, which calculates the credibility of the dual-branch feature maps separately and performs the interactions between branches based on the credibility.}

\subsection{Feature Fusion}

\edita{Feature fusion modules are widely used in various deep learning applications, such as classification, semantic segmentation \cite{ronneberger2015u,zhou2018unet++}, change detection \cite{daudt2018fully,fang2021snunet,chen2022rdpnet}, etc.
Existing modules are primarily for the fusion of multi-scale and multi-modal features \cite{dai2021attentional,feng2021encoder,bandara2022hypertransformer,zhou2022canet,huang2021alignseg}.
They concatenate multiple feature matrices and then generate fusion results through the network layers, or vice versa.
These modules neglect to consider the spatial relationships among bi-temporal data.}
Mohammadian \textit{et al.} \cite{mohammadian2023siamixformer} used the Key, Query, and Value matrices from the self-attention mechanism within Transformer to fuse bi-temporal features and proposed SiamxFormer.
The fusion result is obtained by a temporal transformer.
However, the performance of SiamxFormer is not good enough.

\edita{Additionally, within CD, there exists a simple feature fusion method which directly subtracts bi-temporal data \cite{dai2021attentional,9491802}.
This straightforward method has also achieved good fusion results.
However, it causes the features of unchanged areas to approach zero, which leads to information loss and proves detrimental to the subsequent learning processes.}

\edita{In this paper, we introduce a novel feature fusion module that leverages the bi-temporal spatial relationships of bi-temporal remote sensing data, while avoiding information loss and taking into account the change modes.}

\section{Methodology}

In this section, we first introduce the perception and interaction module, perceiving the spatial relationships among bi-temporal inputs.
Then, we propose the Patch-Mode joint fusion module to avoid information loss and improve the expressive capability of the fusion features.
At last, we present the architecture of the proposed SRC-Net.

\subsection{Perception and Interaction Module}

\begin{figure}[ht]
\centering
\subfloat[]{
    \includegraphics[width= 0.22\linewidth]{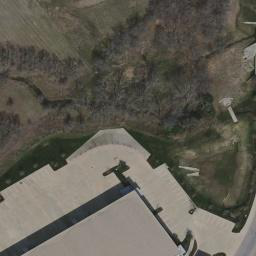}
}
\subfloat[]{
    \includegraphics[width= 0.22\linewidth]{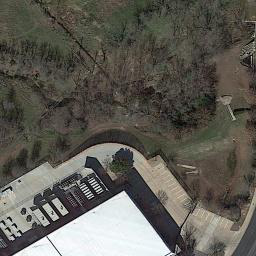}
}
\subfloat[]{
    \includegraphics[width= 0.22\linewidth]{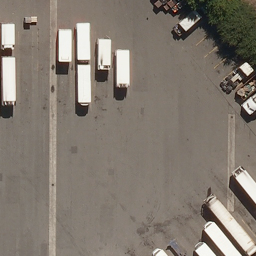}
}
\subfloat[]{
    \includegraphics[width= 0.22\linewidth]{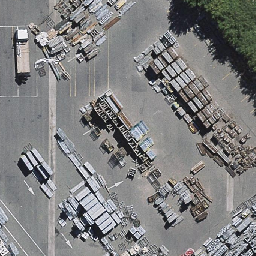}
}
\caption{Original input bi-temporal remote sensing images. (a) and (b) LEVIR-CD dataset, (c) and (d) WHU dataset.}
\label{cpm1}
\end{figure}

Change detection is the analysis of changes in land cover information, facing various challenges and situations.
At times, factors such as weather conditions and lighting can impact the network's capacity to extract features and analyze the land cover information.
In this context, the utilization of bi-temporal comparisons can greatly facilitate the recognition of land cover information.
\edita{As the RGB images shown in Fig. \ref{cpm1}}, the factory buildings at the bottom of Fig. \ref{cpm1}(a) are similar to the nearby roads, making them susceptible to being mistaken for open space rather than industrial structures.
The trucks situated in the top left corner of Fig. \ref{cpm1}(c) are similar to rest areas or service stations which are usually found on the roadside.
\edita{The perception and interaction could help avoid misinterpreting the factory buildings as open space or the trucks as service stations.
This example shows the importance of perception and interaction between branches in CD.}

Therefore, we proposed the perception and interaction module.
For the bi-temporal feature maps extracted by the siamese network, we define their credibility matrices separately, which refer to the probability that the feature map extracted from a single temporal instance is reliable.
When the feature from a single temporal instance is deemed reliable, no adjustments would be made.
However, if the feature is deemed unreliable, we would employ the feature from the other temporal instance to adjust it.
To be more precise, if the other feature is deemed reliable, we would utilize it as the output; otherwise, we would utilize the average of the bi-temporal features as the output.
Drawing inspiration from the self-attention mechanism within Transformer \cite{vaswani2017attention}, wherein the original input computes the Key, Query, and Value matrices directly via neurons, we similarly process bi-temporal feature maps via neurons to acquire their credibility matrices separately.
In practice, \editb{the credibility matrices can be viewed as the probability that the extracted features are correct, by which we can calculate the mathematical expectation of the correct features, and thus obtain the feature interaction results.}

Specifically, when the inputs are two temporal instances, $t_1$ and $t_2$, we first calculate their credibilities separately, denoted as $P_1$ and $P_2$.
Let ${t_1}^{'}$ and  ${t_2}^{'}$ represent the output of this module.
When $t_1$ is deemed reliable, ${t_1}^{'}=t_1$.
When $t_1$ is deemed unreliable and $t_2$ is deemed reliable, ${t_1}^{'}=t_2$.
When both $t_1$ and $t_2$ is deemed unreliable, ${t_1}^{'}=\frac{t_1+t_2}{2}$.
Let $T_1$ represent the event where $t_1$ is deemed reliable, and $\overline{T_1}$ represent the event where $t_1$ is deemed unreliable. The same applies to $T_2$ and $\overline{T_2}$.
The mathematical expectation of ${t_1}^{'}$ can be formulated as:
\begin{equation}
\begin{split}
E\left({t_1}^{'}\right)=&\sum_i \left({t_1}^{'}_{(i)}P_{(i)}\right)\\
=&E\left({t_1}^{'} \mid T_1\right)P\left(T_1\right)+E\left({t_1}^{'} \mid \overline{T_1},T_2\right)P\left(\overline{T_1},T_2\right)\\
&+E\left({t_1}^{'} \mid \overline{T_1},\overline{T_2}\right)P\left(\overline{T_1},\overline{T_2}\right)\\
=&t_1P_1+t_2\left(1-P_1\right)P_2+\frac{t_1+t_2}{2}\left(1-P_1\right)\left(1-P_2\right)
\end{split}
\label{pim1}
\end{equation}
where $i \in \{1,2,3\}$ denotes three potential situations: $T_1$, $\overline{T_1}T_2$ and $\overline{T_1}\overline{T_2}$,
$E\left({t_1}^{'} \mid T_1\right)$ denotes the mathematical expectation of ${t_1}^{'}$ when $t_1$ is deemed reliable, and $P\left(T_1\right)$ denotes the possibility of $T_1$.
The same applies to $E\left({t_1}^{'} \mid \overline{T_1},T_2\right)$, $E\left({t_1}^{'} \mid \overline{T_1},\overline{T_2}\right)$, $P\left(\overline{T_1},T_2\right)$ and $P\left(\overline{T_1},\overline{T_2}\right)$.

Above, we consider $t_1$ and $t_2$ as singular variables.
However, in reality, $t_1$ and $t_2$ represent matrices as feature maps: $t_1, t_2 \in {\mathbb R}^{B\times C\times H\times W}$, where $B$, $C$, $H$ and $W$ represent the batch size, channels, height, and width of the feature map.
The credibilities of the feature maps \{$P_1, P_2$\} are also two matrices: $P_1, P_2 \in {\mathbb R}^{B\times C\times H\times W}$.
The multiplication in (\ref{pim1}) employs the Hadamard product $\odot$ \cite{horn1990hadamard}.
So, the mathematical expectation of ${t_1}^{'}$ is formulated as:
\begin{equation}
\begin{split}
E\left({t_1}^{'}\right)=&t_1\odot P_1+t_2\odot\left(1-P_1\right)\odot P_2\\
&+\frac{t_1+t_2}{2}\odot\left(1-P_1\right)\odot\left(1-P_2\right)
\end{split}
\end{equation}
Similarly, the mathematical expectation of ${t_2}^{'}$ can be formulated as:
\begin{equation}
\begin{split}
E\left({t_2}^{'}\right)=&t_2\odot P_2+t_1\odot\left(1-P_2\right)\odot P_1\\
&+\frac{t_1+t_2}{2}\odot\left(1-P_2\right)\odot\left(1-P_1\right)
\end{split}
\end{equation}

We name this feature interaction as the perception and interaction module.
The credibility matrices are inferred from the input instances by linear layers, which can be formulated as follows:
\begin{equation}
P_i={\rm sigmoid}\left({\rm Linear}(C,C)(t_i)\right)
\end{equation}
where ${\rm Linear}(C, C)(\cdot)$ refers to a linear layer with $C$ as input and output dimensions.
The function ${\rm sigmoid}(\cdot)$ maps a real number to the interval of $(0,1)$.

\begin{figure}[ht]
\centering
\includegraphics[width= 0.6\linewidth]{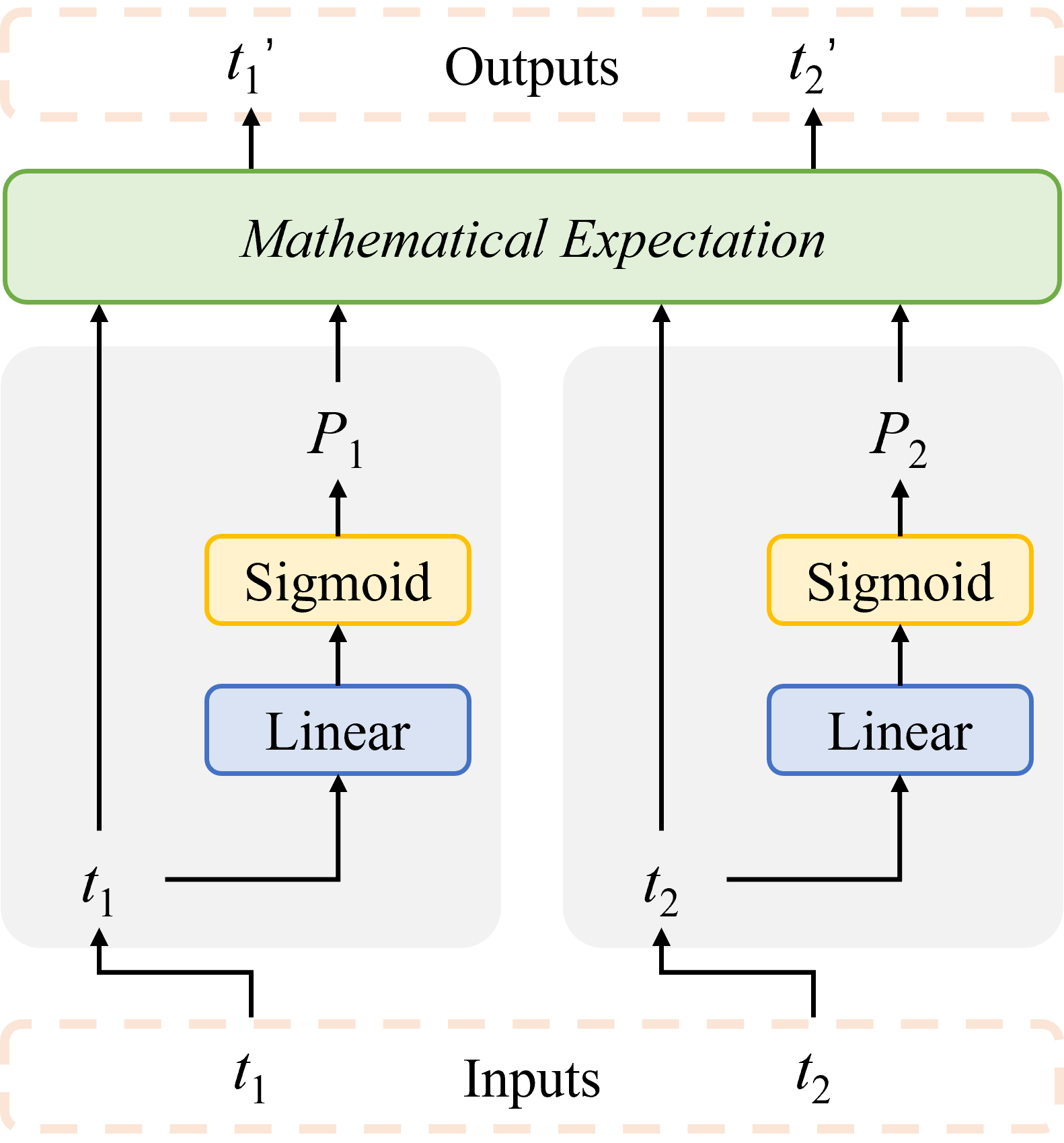}
\caption{The structure of perception and interaction module.}
\label{pimstructure}
\end{figure}

The structure of the perception and interaction module is shown in Fig. \ref{pimstructure}.
Our module takes into account the bi-temporal spatial relationships in remote sensing CD, enabling the siamese network to engage in dual-branch communication during the process of feature extraction. 
This enhances the network's ability to extract features from bi-temporal remote sensing images.

\subsection{Patch-Mode joint Feature Fusion Module}

One crucial task within CD is the fusion of feature maps extracted by siamese networks from different temporal instances.
Existing methods commonly employ concatenation or subtraction of bi-temporal features.
As mentioned previously, existing methods inevitably result in information loss, and a lack of consideration for various situations of change.
It is worth delving into bi-temporal feature fusion to serve the subsequent prediction.
The fusion module should take into account various situations of change and the bi-temporal spatial relationships, thereby obtaining more expressive and robust fusion features.

\begin{figure*}[ht]
\centering
\includegraphics[width= \linewidth]{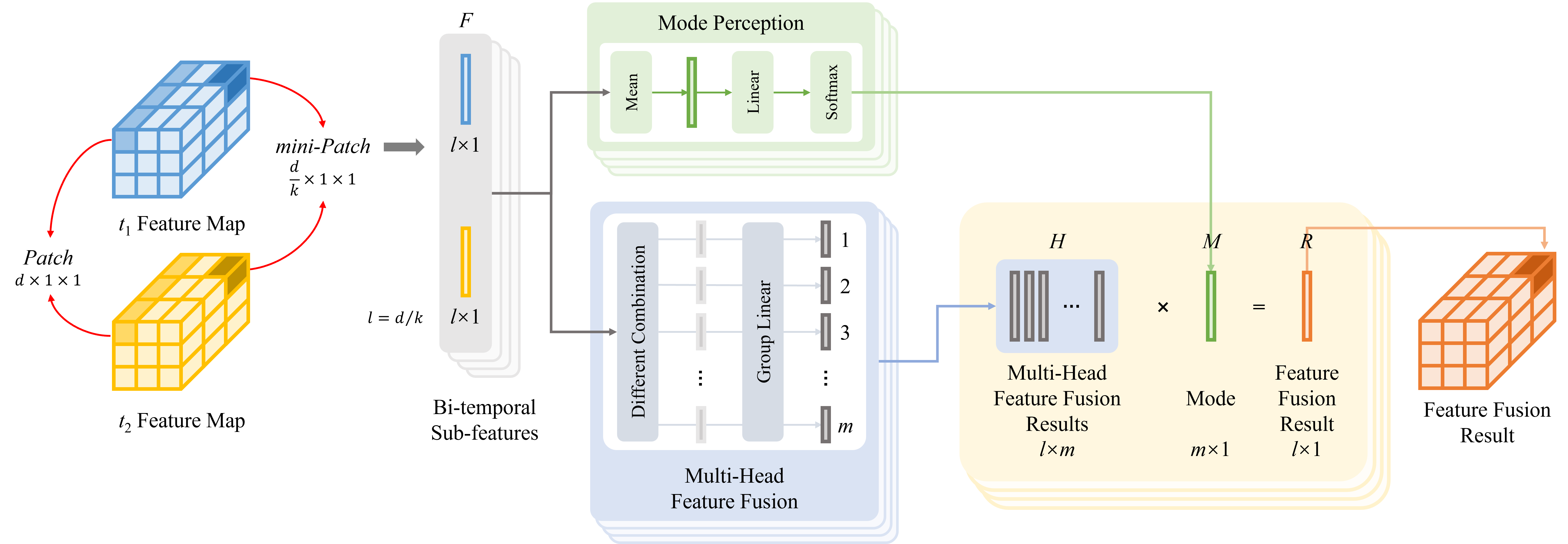}
\caption{The structure of the Patch-Mode joint feature fusion module.}
\label{ffmS}
\end{figure*}

Therefore, we propose the Patch-Mode joint feature fusion module, inspired by the methods of handling time-varying signals in the field of signals and systems.
In the realm of signal and system analysis, when examining a time-varying signal, it is common practice to extract the fundamental frequency or baseline of a time-varying signal first, which provides a macroscopic description of the signal's overall behavior.
Derived from this, we can determine the mode of the signals, subsequently allowing for the selection of an optimal approach for signal processing.
A similar idea should also be adopted for remote sensing CD.
Different feature fusion heads should be used to fuse the bi-temporal feature maps corresponding to different modes.
The mean of bi-temporal features can be regarded as the baseline, around which the features of preceding and subsequent temporal instances exhibit variations.
Based on the baseline, we deduce the probabilities of various change modes, and obtain fusion features corresponding to these distinct modes through the specialized fusion heads.
Subsequently, the final fusion result is ascertained based on mathematical expectation.
Specifically, the structure of the Patch-Mode joint feature fusion module is shown in Fig. \ref{ffmS}.

\subsubsection{Bi-temporal Sub-features}

We partition the input image into patches for learning at the beginning, so the size of the feature map extracted by the siamese network can be equivalently represented as $d\times h\times w$, where $h$ and $w$ denote the height and width of the input feature map, $d$ denotes the depth.
The size of each patch is $d\times 1\times 1$.
Considering that the modes within each patch may also vary, each patch is further subdivided into $k$ mini-patches.
The size of the mini-patches is $\frac{d}{k}\times 1\times 1$.
This allows for a more detailed determination of the modes within each patch.
A mini-patch equivalently refers to a vector of dimensions $l\times 1$, where $l=d/k$.
The mini-patches from the same position in the bi-temporal feature maps are combined to form multiple sets of bi-temporal sub-features.

\subsubsection{Mode Perception}

A mode perception mechanism is constructed to determine the mode of the bi-temporal sub-features.
As mentioned above, we consider the mean of the bi-temporal sub-features as the baseline and utilize a linear layer followed by a softmax layer to infer the probabilities of the different modes for each sub-feature.
The mode perception mechanism can be formulated as
\begin{equation}
M_i={\rm Softmax}\left({\rm Linear}(l,m)({\rm mean}(F_i))\right)
\end{equation}
where $F_i=\left\{F_{i,1},F_{i,2}\right\}$ represents the \textit{i}-th bi-temporal sub-feature, $F_{i,1}, F_{i,2}\in {\mathbb R}^{l\times 1}$ represent the \textit{i}-th sub-feature of $t_1$ and $t_2$ feature map, $M_i \in {\mathbb R}^{m\times 1}$ represents the \textit{i}-th probability matrix of the different modes, $m$ represents the total number of modes.
${\rm mean}(\cdot)$ refers to the operation of taking the mean value.
${\rm Linear}(l, m)(\cdot)$ refers to a linear layer with $l$ as input dimension and $m$ as output dimension.
${\rm Softmax}(\cdot)$ refers to the softmax function.

\subsubsection{Multi-Head Feature Fusion}

In correspondence with the various modes that exist for sub-features, we design a multi-head feature fusion mechanism.
The inputs for different heads are obtained through linear combinations of the sub-feature.
Different combinations are processed through a group linear layer to obtain the fusion results of different modes.
The number of groups in the group linear layer is equal to the number of modes, implying that each mode has its dedicated linear layer.
This arrangement allows for generating unique feature fusion results for each mode.
The combination of bi-temporal sub-features can be formulated as
\begin{equation}
C_{i,j}=\alpha_jF_{i,1}+\beta_jF_{i,2}
\end{equation}
where $\alpha_j,\beta_j$ represent the \textit{j}-th coefficients for the linear combination of $F_{i,1}$ and $F_{i,2}$, $C_{i,j}\in {\mathbb R}^{l\times 1}$ represents the \textit{j}-th linear combination, $j \in \{1,2,...,m\}$ of $F_i$.
The multi-head feature fusion results can be formulated as
\begin{equation}
H_{i,j}={\rm Linear}_{(j)}(l,l)(C_{i,j})
\end{equation}
where ${\rm Linear}_{(j)}(l, l)(\cdot)$ refers to the \textit{j}-th linear layer with $l$ as input and output dimension in the group linear layer.
$H_{i,j}\in {\mathbb R}^{l\times 1}$ represents the feature fusion result of the \textit{j}-th head.
\begin{equation}
\begin{split}
H_i&={\rm Concat}\left(H_{i,1},H_{i,2},...,H_{i,m}\right)\\
&=\left[H_{i,1},H_{i,2},...,H_{i,m}\right]
\end{split}
\end{equation}
where ${\rm Concat}(\cdot)$ refers to concatenating matrices along the second dimension, $H_i\in {\mathbb R}^{l\times m}$ represents the multi-head feature fusion results.

Consequently, we obtain the probabilities of different modes for each sub-feature, and obtain the fusion results by utilizing the feature fusion head specific to each mode.
The final feature fusion result of a sub-feature can be obtained through matrix multiplication, which can be formulated as
\begin{equation}
R_i=H_i\times M_i
\end{equation}
where $R_i\in {\mathbb R}^{l\times 1}$ represents the final feature fusion result of the sub-feature $F_i$.
The fusion result of $t_1$ and $t_2$ feature maps is obtained by combining the results of every bi-temporal sub-feature.

The proposed Patch-Mode joint Feature Fusion Module effectively avoids information loss, takes into account the different change modes, and utilizes the bi-temporal spatial relationships of the input instances.
In the experiment, the number of mini-patches in each patch $k$ is set to 16, and the number of feature fusion heads $m$ is also set to 4.

\subsection{Network Architecture}

\begin{figure*}[ht]
\centering
\includegraphics[width= \linewidth]{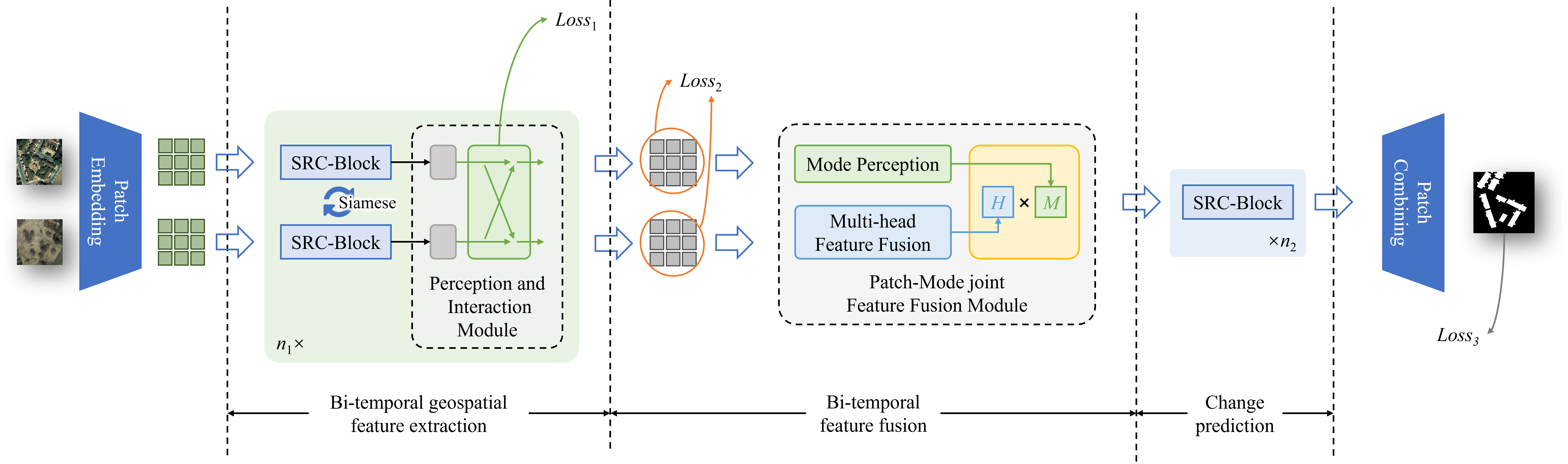}
\caption{Architecture of the proposed SRC-Net.}
\label{STC}
\end{figure*}

The overall architecture of the proposed SRC-Net is shown in Fig. \ref{STC}.
The network backbone is similar to RDP-Net, but is enhanced by employing siamese networks to extract features separately from bi-temporal inputs.
And we augment the supervision of network learning.
Our network primarily consists of three stages: bi-temporal geospatial feature extraction stage, bi-temporal feature fusion stage, and change prediction stage, complemented by two additional layers: patch embedding layer and patch combining layer.
In addition to supervising the final prediction, we have introduced supervision for the features extracted by the bi-temporal geospatial feature extraction stage.
\edita{The bi-temporal features obtained in the feature extraction stage possess the capacity to characterize ground object information.
By analyzing these features, we can identify the types of ground objects at different times and initially determine the change areas.
The comparison of the initial result with the ground truth ($Loss_2$) supervises the feature extraction stage.}

Specifically, the patch embedding layer is employed to partition the input image into multiple patches based on their respective regions.
In order to reduce the loss of information and maintain the intricate local details, the patch embedding layer with patch size ($p \times p$) and input image size ($C \times H \times W$) can be implemented as two convolutions (${\rm GenPatch(\cdot)}$ and ${\rm GenPatch_2(\cdot)}$).
The input channels of the first convolution ${\rm GenPatch(\cdot)}$ are $c_{in}$, output channels are $c/4$, kernel size is $4$, and stride is $4$.
The input channels of the second convolution ${\rm GenPatch_2(\cdot)}$ are $c/4$, output channels are $c$, kernel size is $p/4$, and stride is $p/4$.
The layer can be formulated as
\begin{equation}
X_{patch}={\rm GenPatch_2} \left( {\rm BN}( {\rm GenPatch} (X_{input}) ) \right)
\end{equation}
where ${\rm BN}()$ represents Batch Normalization \cite{ioffe2015batch}, \editb{$X_{patch}$ represents the patch embedding feature maps, and $X_{input}$ represents the input image.}

\begin{figure}[ht]
\centering
\includegraphics[width= 0.6\linewidth]{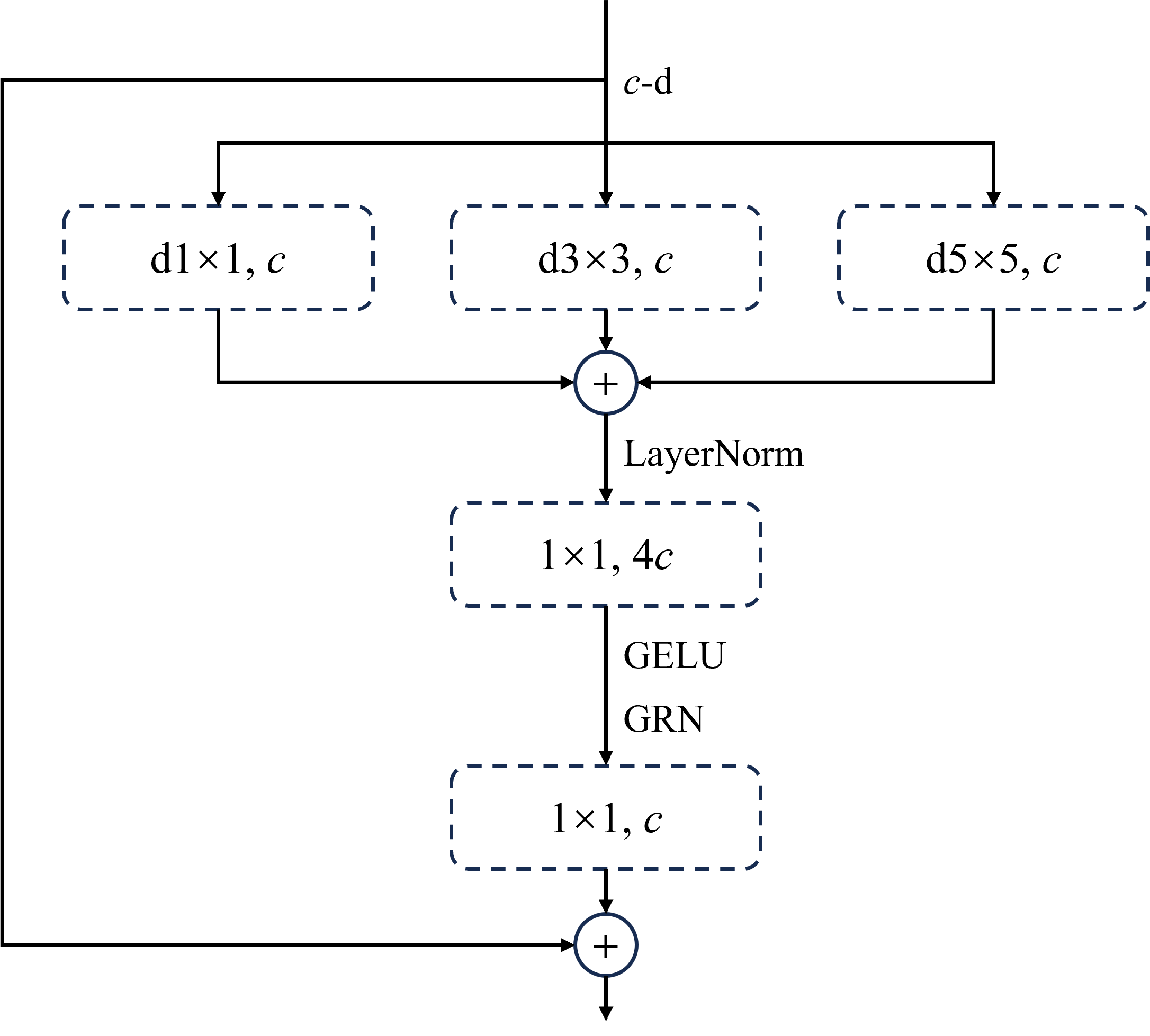}
\caption{SRC-Block.}
\label{STCblock}
\end{figure}

The SRC-Block is used as the backbone of our network.
As shown in Fig. \ref{STCblock}, it consists of a depthwise convolutional layer followed by two pointwise convolutional layers.
The depthwise layer consists of three parallelized group convolutional layers, which are used to explore the information between different patches.
Each of the three group convolutional layers has $c$ input and output channels, with $c$ groups. The kernel sizes for these layers are 1, 3, and 5, respectively.
The two pointwise convolutional layers are used to explore the information of each patch, where we explore the inverted bottleneck design.
The kernel sizes for these convolutional layers are 1.
The input channels and output channels for these layers are $\left(c, 4c\right)$, and $\left(4c, c\right)$ respectively.
The residual connection serves to maintain a primary focus on the details inside a patch when the network explores the context information between patches.
LayerNorm represents layer normalization \cite{ba2016layer}, GELU represents Gaussian error linear unit \cite{hendrycks2016gaussian}, and GRN represents the global response normalization layer \cite{woo2023convnext}.

Bi-temporal geospatial feature extraction stage:
\edita{This stage employs siamese networks to extract features respectively from bi-temporal inputs.
Each siamese branch contains $n_1$ blocks, with each block consisting of a siamese SRC-Block and a perception and interaction module.
The SRC-Blocks on each branch of the Siamese networks share network parameters.}
These $n_1$ blocks can extract features from bi-temporal inputs, harnessing the bi-temporal spatial relationships and engaging in dual-branch communication.

Bi-temporal feature fusion stage:
This stage uses the Patch-Mode joint feature fusion module, which takes into account the varying situations of changes, preserves bi-temporal spatial relationships, and avoids information loss.

Change prediction stage:
This stage contains $n_2$ SRC-Blocks to predict the change results from the fusion feature maps.

The patch combining layer is utilized to merge patches based on their respective regions in order to generate a pixel-level classification feature map.
The region composition layer can be implemented as a convolution-transpose and a convolution.
The input channels of the convolution-transpose ${\rm PatchUp(\cdot)}$ are $c$, output channels are 32, the kernel size is $p$, and stride is $p$.
The input channels of the convolution ${\rm Final(\cdot)}$ are 32, output channels are $out\_ch$, and the kernel size is 1.
The layer can be formulated as
\begin{equation}
Y_{output}={\rm Final} \left( {\rm GELU} \left( {\rm BN}( {\rm PatchUp} (Y_{patch}) ) \right) \right)
\end{equation}
\editb{where $Y_{output}$ represents the predicted results, and $Y_{patch}$ represents the input patch embedding feature maps.}

The previous paragraphs delineated the architecture of the different stages within the network.
To facilitate a more robust learning process, we have implemented three sets of loss functions.
These loss functions supervise the network's training together.
The overall loss function can be defined as
\begin{equation}
Loss=Loss_1+Loss_2+Loss_3
\end{equation}

$Loss_1$ primarily imposes constraints on the training of the perception and interaction module in the bi-temporal geospatial feature extraction stage, ensuring that the module can better retain pertinent information.
Specifically, a singular temporal instance along with its noisy signal is fed into the module.
\editb{The credibilities calculated for the noisy signal should be lower, and the credibilities of the original signal should be higher, resulting in two outputs that are closer to the original signal.}
We calculate the disparity between the module's output signal and the original data as the loss, \edita{which can be formulated as
\begin{equation}
Loss_1=\left|\left|S_{out}-S_{ori}\right|\right|_2
\end{equation}
where $S_{out}$ represents the module's output signal, $S_{ori}$ represents the original data.}

$Loss_2$ and $Loss_3$ both describe the disparity between the prediction and the ground truth.
$Loss_2$ primarily examines the feature extraction capability of the feature extraction network.
Utilizing a convolution-transpose and a softmax layer, the likelihood of land cover types for each point can be inferred.
Then, we can calculate the probability of change at each point by employing Bayes' formula, thereby computing the loss, which enables effective supervision of the feature extraction.
$Loss_3$ is employed to calculate the disparity between the final prediction and the ground truth.
In our network, we use a hybrid loss function as $Loss_2$ and $Loss_3$, which can be defined as ${\mathcal L}$
\begin{equation}
{\mathcal L}=\frac{1}{\sigma_1^2}{\mathcal L}_{\rm focal}+\frac{1}{\sigma_2^2}{\mathcal L}_{\rm dice}+\frac{1}{\sigma_3^2}{\mathcal L}_{\rm edge}+\log \sigma_1\sigma_2\sigma_3
\end{equation}
where ${\mathcal L}_{\rm focal}$ represents Focal loss \cite{lin2017focal}, ${\mathcal L}_{\rm dice}$ represents Dice loss \cite{milletari2016v} and ${\mathcal L}_{\rm edge}$ represents Edge loss \cite{chen2022rdpnet}.
$\sigma_1$, $\sigma_2$ and $\sigma_3$ represent the learnable relative weight of the losses ${\mathcal L}_{\rm focal}$, ${\mathcal L}_{\rm dice}$ and ${\mathcal L}_{\rm edge}$, according to a principled approach to multi-task deep learning proposed by Alex \textit{et al.} \cite{kendall2018multi}.
Focal loss and dice loss primarily address issues related to class imbalance.
Edge loss is primarily concerned with capturing details, particularly in boundary regions.
\editb{The formula of Edge Loss is
\begin{equation}
{\mathcal L}_{\rm edge}=- w_{\textsf{edge}} \log (p_t)
\end{equation}
where $w_{\textsf{edge}}$ represents the edge weight of point, $p_t$ represents the probability of correct classification.}

SRC-Net source code is available at \url{https://github.com/Chnja/SRCNet}.

\section{Experiments and Analysis}

\subsection{Dataset}

The experiment was conducted on two datasets named LEVIR-CD dataset \cite{Chen2020} and WHU Building dataset \cite{ji2018fully}, two of the most common datasets in remote sensing change detection.

\textit{LEVIR-CD} dataset consists of 637 \edita{visible light} image pairs of $1024 \times 1024$ pixels, \edita{which are collected from Google Earth.}
\edita{The spatial resolution is $0.5m$ per pixel.}
We cut each image pair into $256 \times 256$ pixel patches without overlapping and ultimately obtained 7120 training sets and 1024 validation sets.

\textit{WHU Building} dataset consists of two \edita{visible light} aerial images of size $32507 \times 15354$, \edita{which are captured by NZ Aerial Mapping Ltd.}
\edita{The spatial resolution is $0.3m$ per pixel.}
We cut each image pair into $256 \times 256$ pixel patches without overlapping and ultimately obtained 5141 training sets and 2293 validation sets.

\subsection{Implementation Details}

We implemented SRC-Net using the Pytorch framework.
The number of perception and interaction modules in the bi-temporal geospatial feature extraction stage is set to 4, and the number of SRC-Blocks in the change prediction stage is also set to 4.
$c$ is set to 256. 
The learning rate is set to 2e-3 and decays by 0.8 every 20 epochs.
In the training process, the batch size is set to 16, and AdamW \cite{loshchilov2017decoupled} is applied as an optimizer.
We conducted experiments on a single NVIDIA RTX3090 and trained for 300 epochs.
\editb{The comparison methods also use all the same parameters.}

We used three metrics, precision, recall, the F1 score, \editb{Overall Accuracy (OA), and IoU} to quantitatively evaluate the performance of the CD models.

\subsection{Comparison With SOTA Methods}

We compared our method with FC-EF, FC-Siam-conc, FC-Siam-diff \cite{daudt2018fully}, STANet \cite{chen2020spatial}, BIT \cite{9491802}, L-UNet \cite{9352207}, DSIFN \cite{zhang2020deeply}, SNUNet \cite{fang2021snunet}, RDP-Net \cite{chen2022rdpnet}, Changer \cite{fang2023changer}, SiamixFormer \cite{mohammadian2023siamixformer}, BAN \cite{li2024new}, and LightCDNet \cite{xing2023lightcdnet}.
These methods represent deep learning-based approaches in the field of CD.
FC-EF, FC-Siam-conc, and FC-Siam-diff \cite{daudt2018fully} are baseline models and they represent improvements of U-Net \cite{ronneberger2015u}.
STANet \cite{chen2020spatial} incorporates a spatial-temporal attention mechanism to improve the discrimination of deep features within bi-temporal images.
BIT \cite{9491802} utilizes a customized CNN along with a pair of transformer encoders and decoders to effectively address the CD problem.
L-UNet \cite{9352207} represents a combination of convolutional and recurrent approaches, employing fully convolutional LSTM blocks for an end-to-end neural network.
DSIFN \cite{zhang2020deeply}, SNUNet \cite{fang2021snunet}, and LightCDNet \cite{xing2023lightcdnet} are three recently proposed methods, which are effective in obtaining competitive results.
Changer \cite{fang2023changer} proposes a novel general architecture, MetaChanger, which integrates interaction layers within the process of feature extraction.
SiamixFormer \cite{mohammadian2023siamixformer} leverages temporal transformers in feature fusion, enhancing the preservation of large receptive fields generated by transformer encoders.
BAN \cite{li2024new} proposes a universal foundation model-based CD adaptation framework designed to extract foundational model knowledge for change detection purposes.

\begin{figure*}[!ht]
\centering
\subfloat[]{
    \includegraphics[width= 0.1\textwidth]{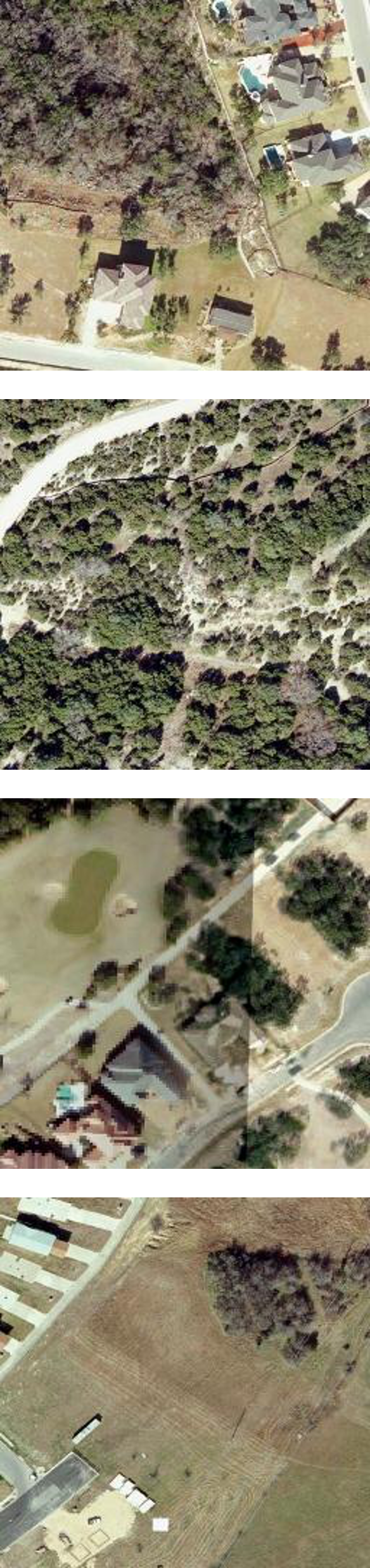}
}
\subfloat[]{
    \includegraphics[width= 0.1\textwidth]{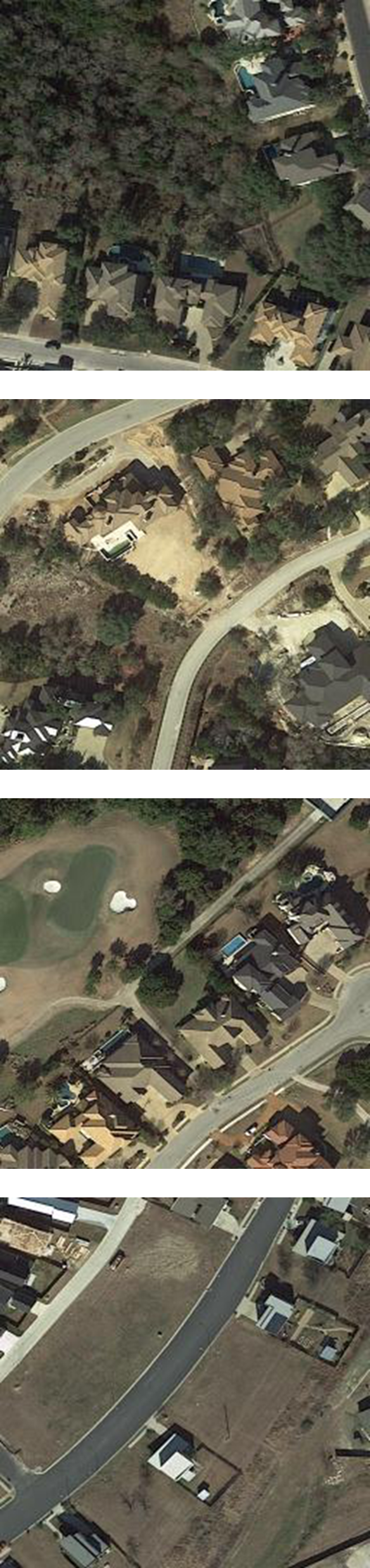}
}
\subfloat[]{
    \includegraphics[width= 0.1\textwidth]{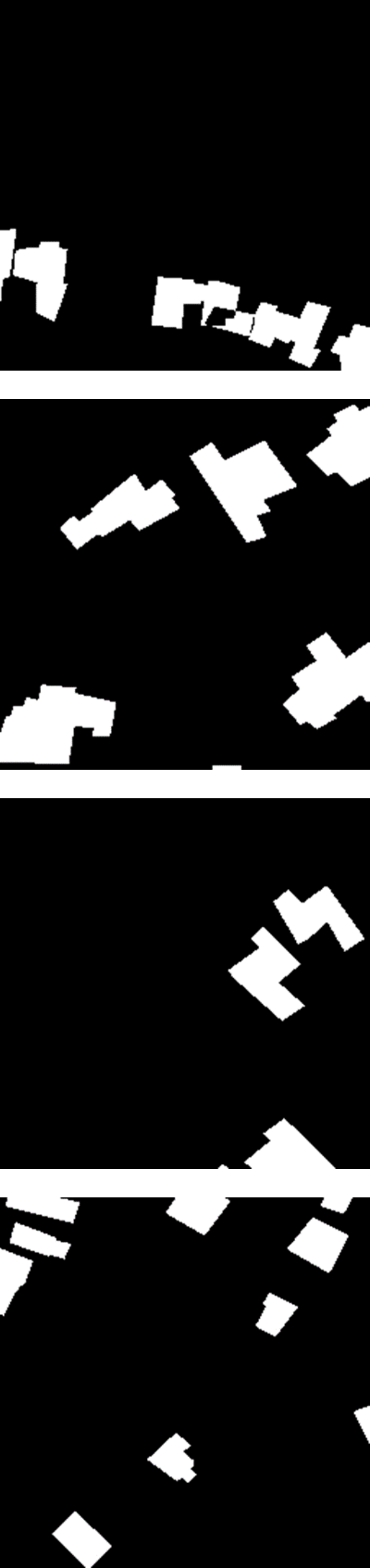}
}
\subfloat[]{
    \includegraphics[width= 0.1\textwidth]{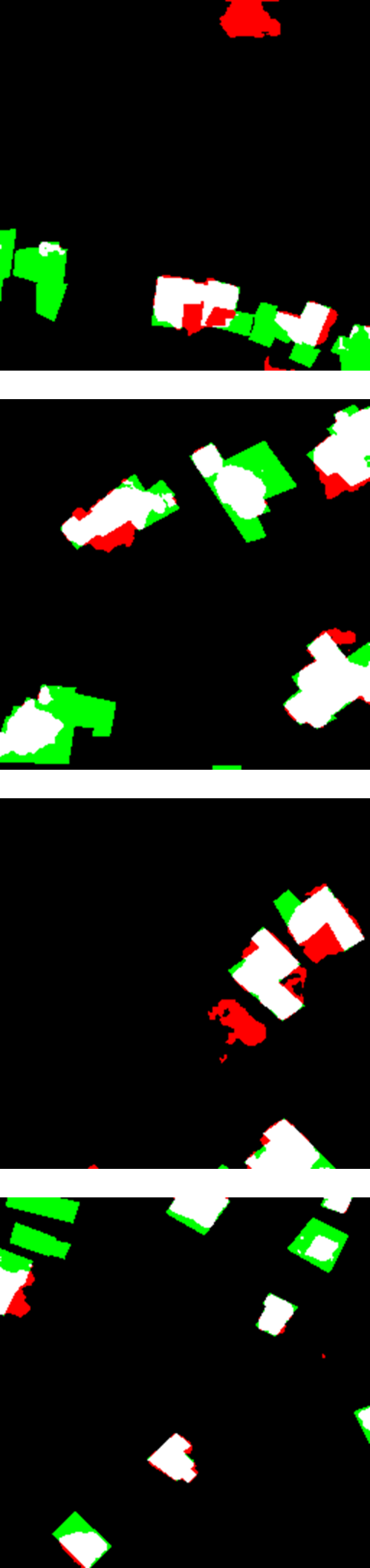}
}
\subfloat[]{
    \includegraphics[width= 0.1\textwidth]{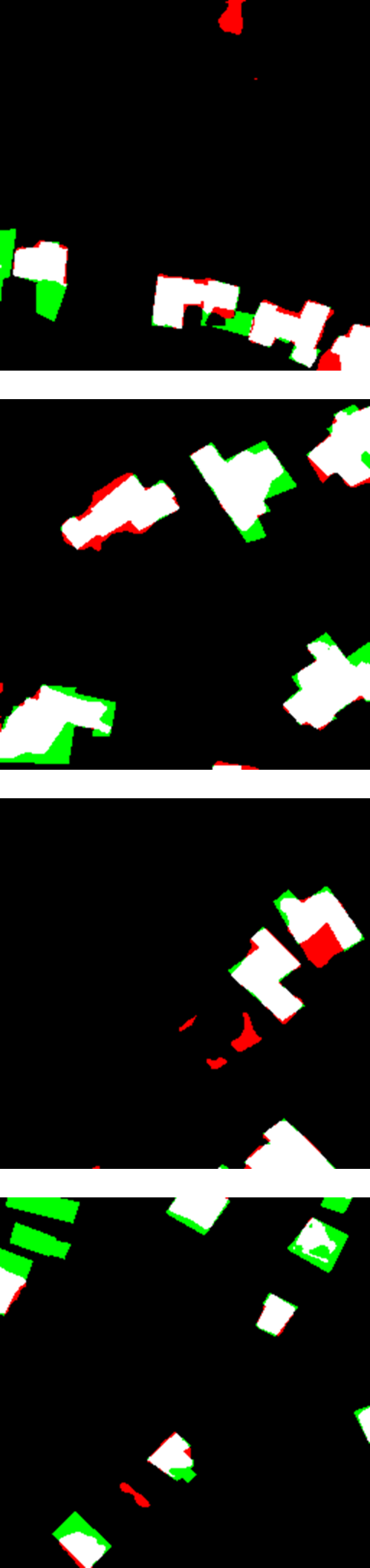}
}
\subfloat[]{
    \includegraphics[width= 0.1\textwidth]{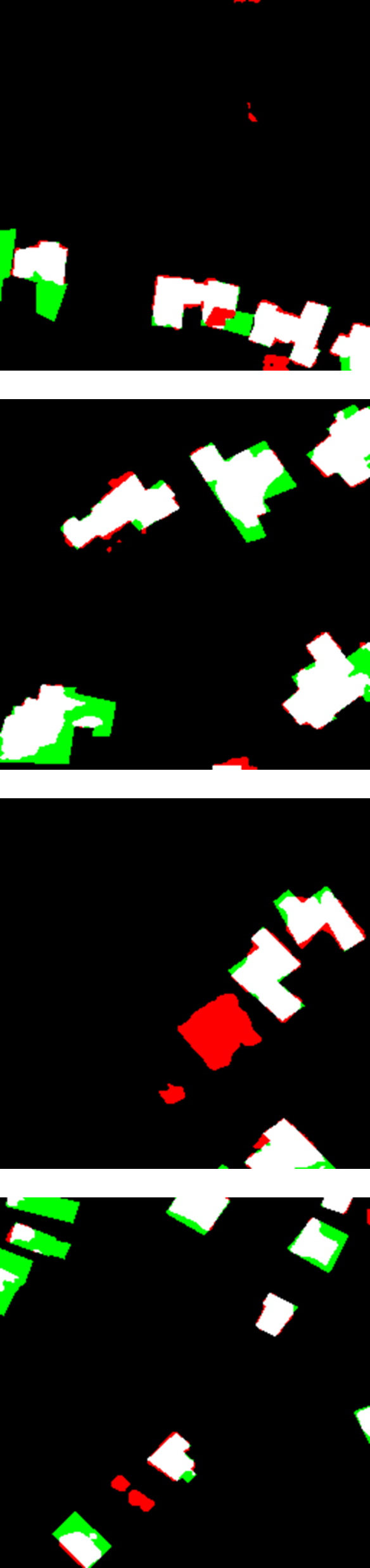}
}
\subfloat[]{
    \includegraphics[width= 0.1\textwidth]{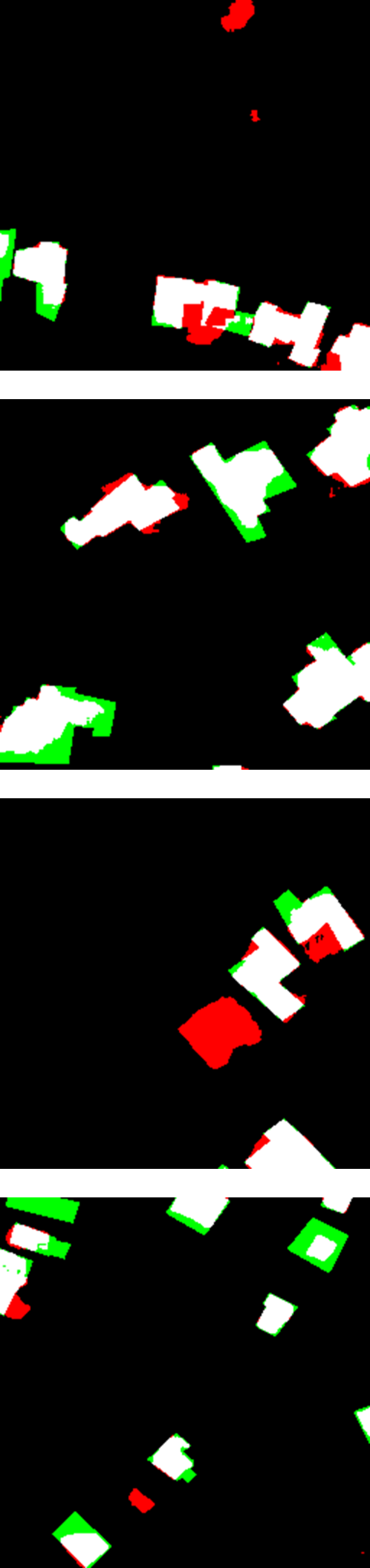}
}
\subfloat[]{
    \includegraphics[width= 0.1\textwidth]{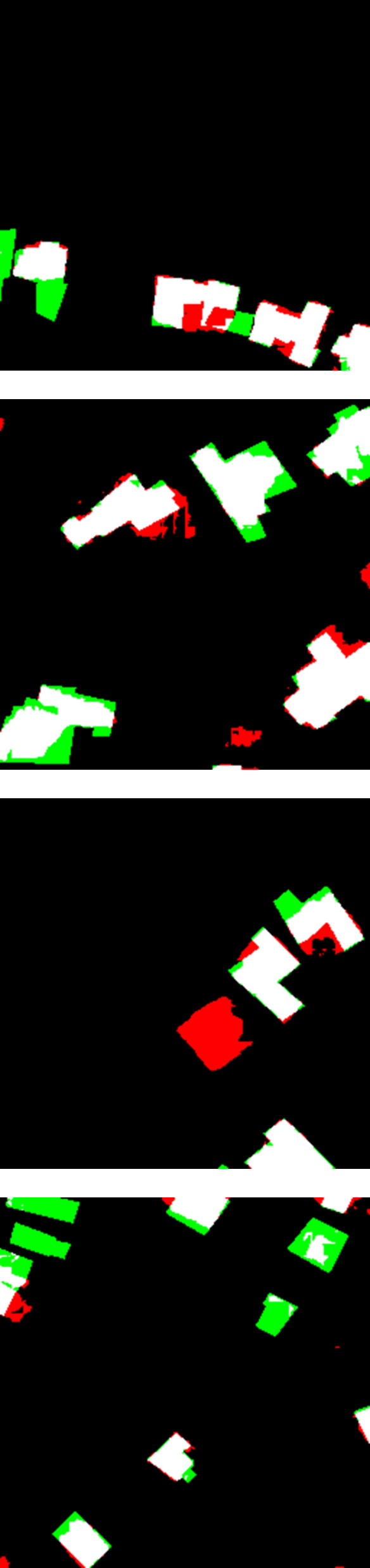}
}
\subfloat[]{
    \includegraphics[width= 0.1\textwidth]{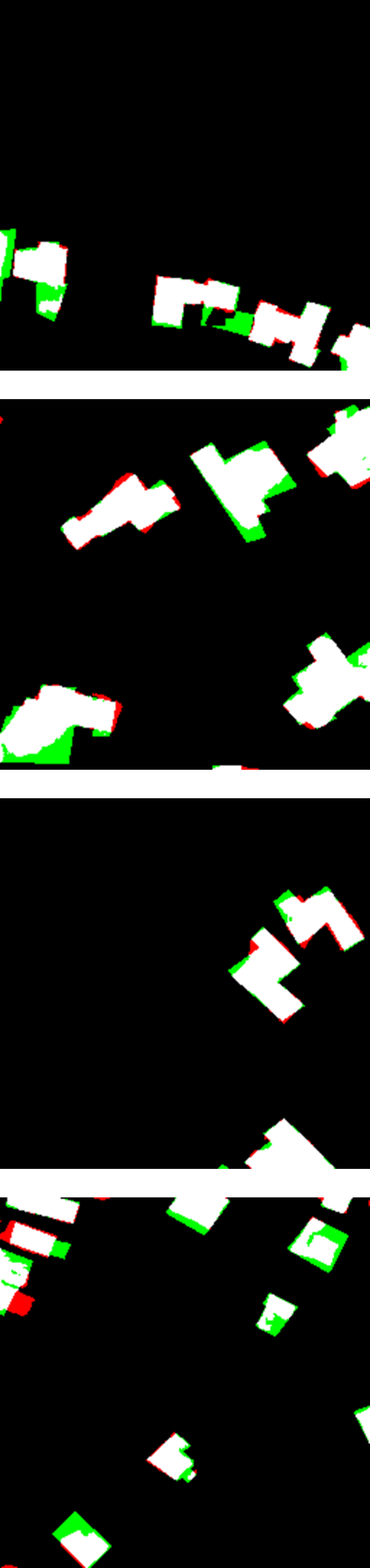}
}

\subfloat[]{
    \includegraphics[width= 0.1\textwidth]{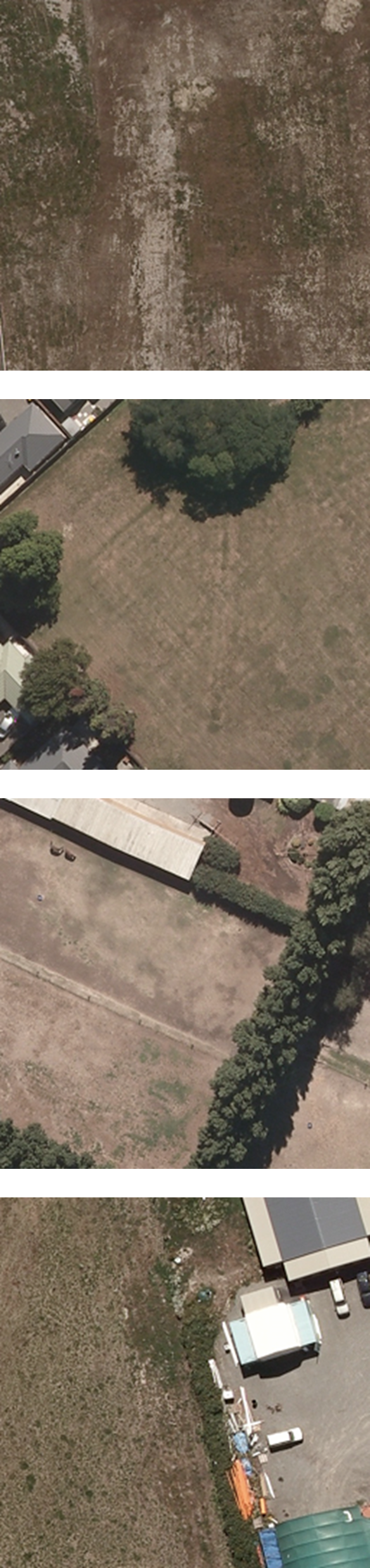}
}
\subfloat[]{
    \includegraphics[width= 0.1\textwidth]{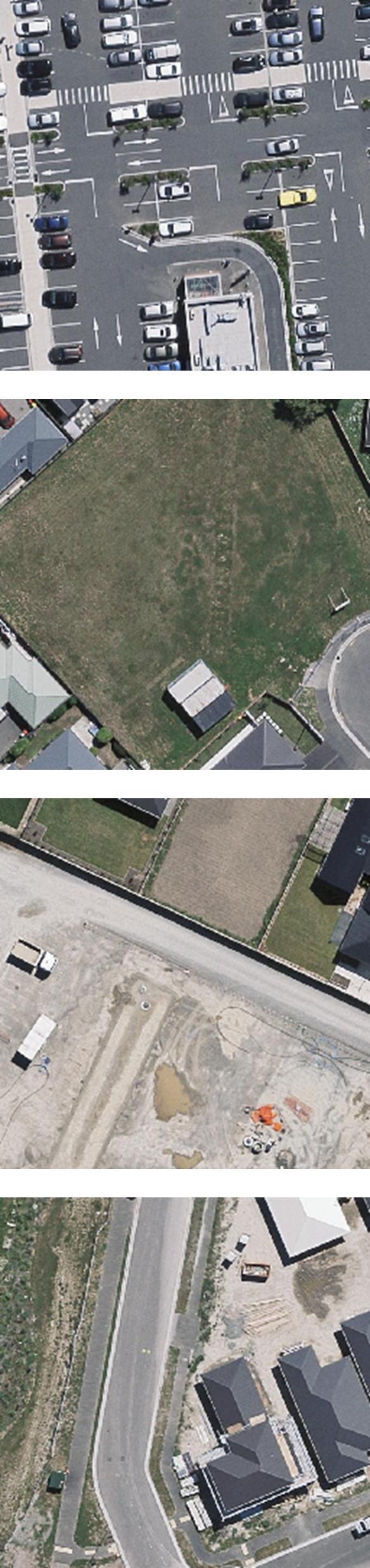}
}
\subfloat[]{
    \includegraphics[width= 0.1\textwidth]{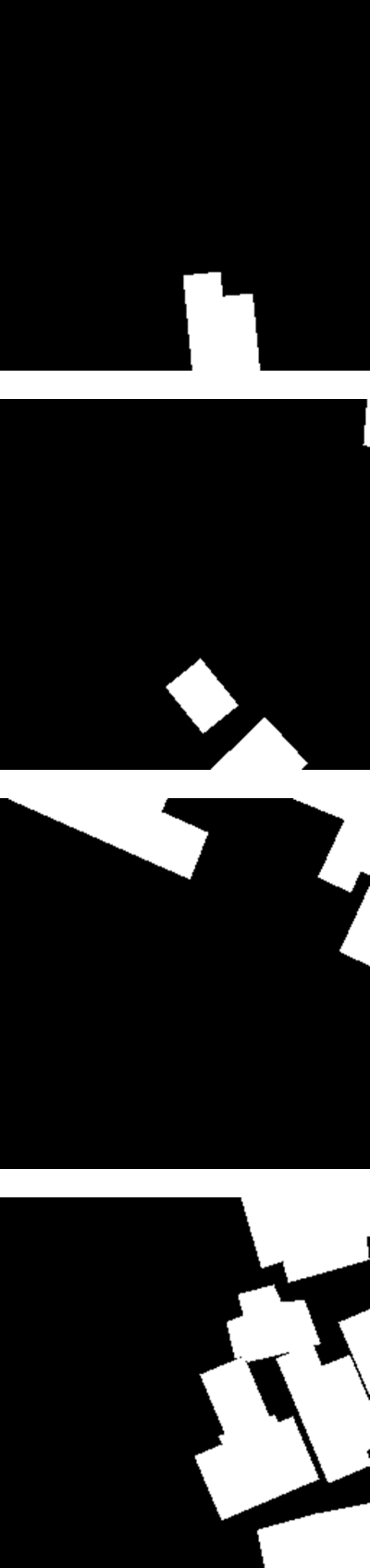}
}
\subfloat[]{
    \includegraphics[width= 0.1\textwidth]{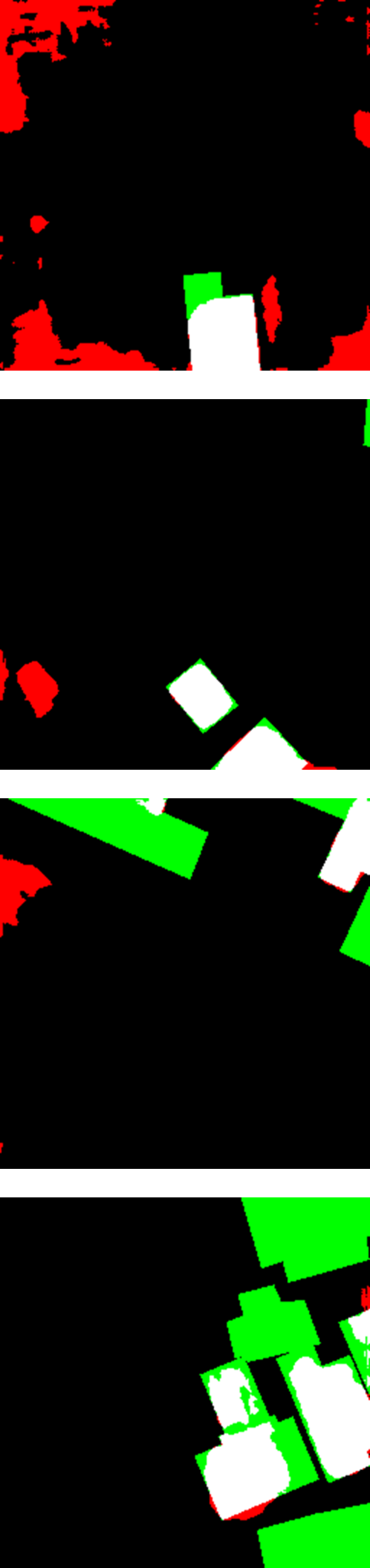}
}
\subfloat[]{
    \includegraphics[width= 0.1\textwidth]{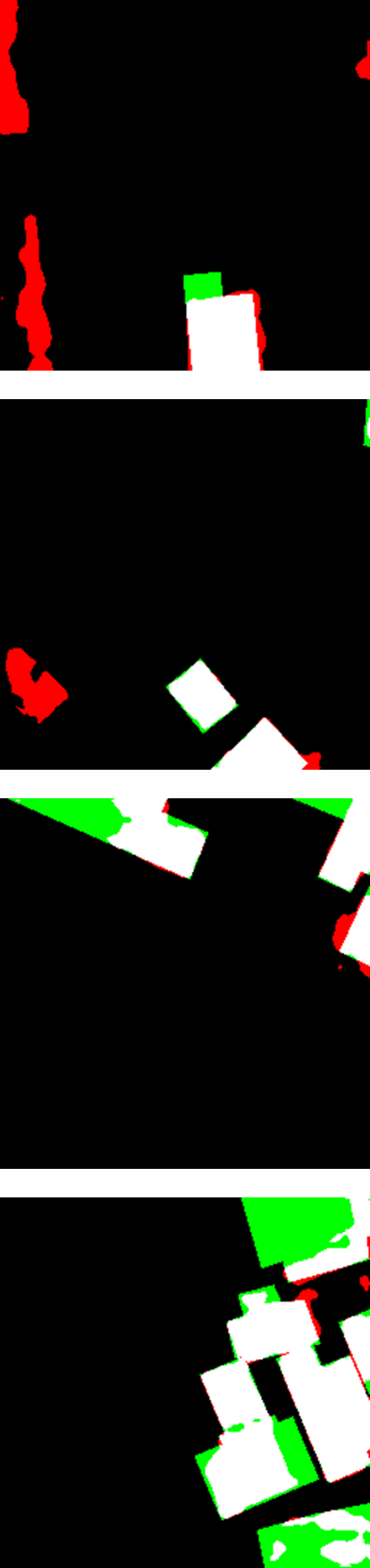}
}
\subfloat[]{
    \includegraphics[width= 0.1\textwidth]{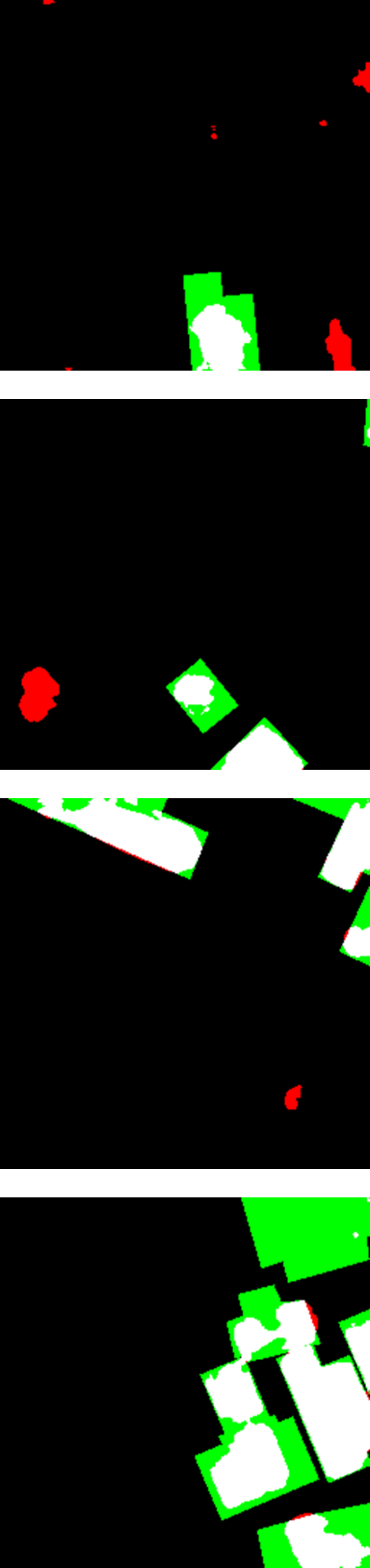}
}
\subfloat[]{
    \includegraphics[width= 0.1\textwidth]{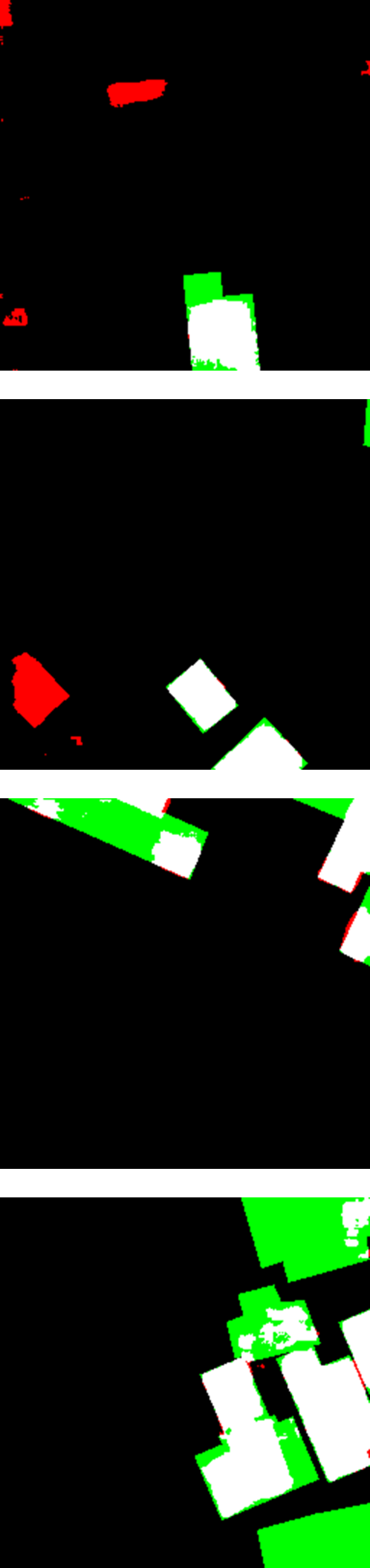}
}
\subfloat[]{
    \includegraphics[width= 0.1\textwidth]{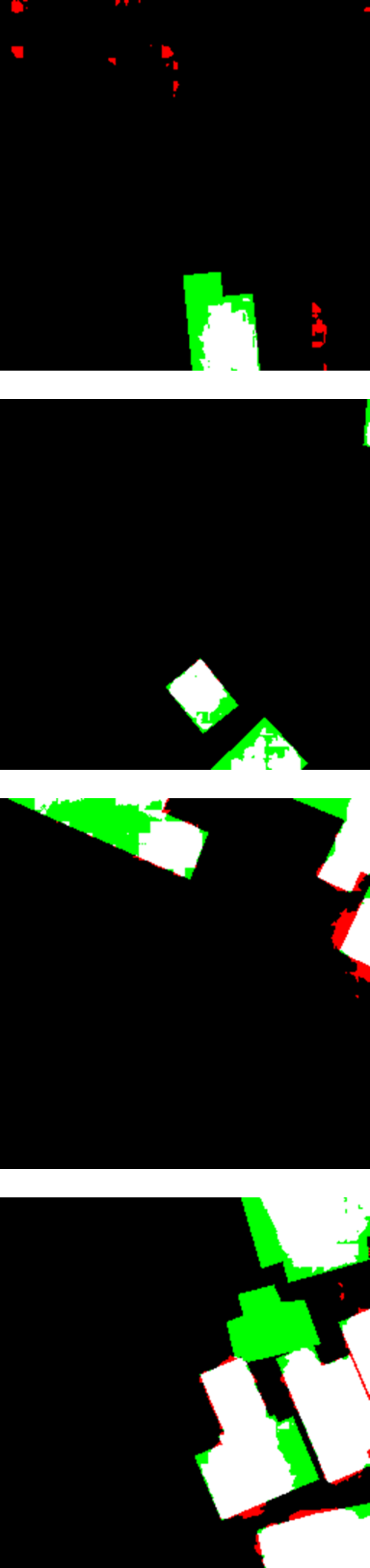}
}
\subfloat[]{
    \includegraphics[width= 0.1\textwidth]{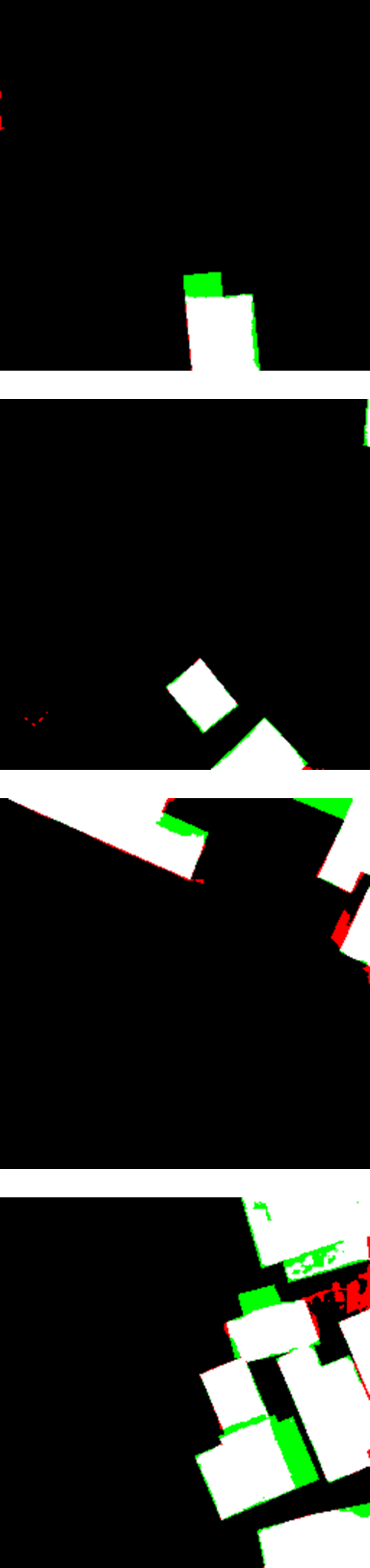}
}
\caption{(a)-(i) are results on the LEVIR-CD dataset.
(j)-(r) are results on the WHU Building dataset.
(a), (b), (j), and (k) are the original images.
(c) and (l) are the ground truth.
The results of (d) (m) FC-Siam-diff,
(e) (n) BIT,
(f) (o) DSIFN,
(g) (p) SNUNet,
(h) (q) RDP-Net,
(i) (r) our SRC-Net.
The false positives and false negatives are indicated in red and green, respectively.
Other colors represent true positives.}
\label{SOTA}
\end{figure*}

\begin{table*}[ht]\small
\caption{Results on LEVIR-CD Dataset and WHU Building Dataset.}
\label{tab:result}
\centering
\begin{tabular}{cccccccccccc}
\hline
\hline
\multirow{2}*{Model} & \multirow{2}*{Params(M)} & \multicolumn{5}{c}{LEVIR-CD dataset} & \multicolumn{5}{c}{WHU Building dataset}\\
\cline{3-12}
 & & Precision & Recall & F1 & \editb{OA} & \editb{IoU} & Precision & Recall & F1 & \editb{OA} & \editb{IoU}\\
\hline
FC-EF & 1.35 & 0.7460 & 0.8826 & 0.8086 & \editb{0.9825} & \editb{0.6787} & 0.7842 & 0.7462 & 0.7647 & \editb{0.9796} & \editb{0.6191}\\
FC-Siam-diff & 1.35 & 0.8876 & 0.8789 & 0.8832 & \editb{0.9902} & \editb{0.7908} & 0.6740 & 0.8194 & 0.7993 & \editb{0.9817} & \editb{0.6656}\\
FC-Siam-conc & 1.55 & 0.8904 & 0.8899 & 0.8902 & \editb{0.9908} & \editb{0.8020} & 0.6106 & 0.8668 & 0.7165 & \editb{0.9695} & \editb{0.5582}\\
STANet & 16.97 & \textbf{0.9338} & 0.8658 & 0.8985 & \editb{0.9910} & \editb{0.8157} & 0.9190 & 0.8827 & 0.9005 & \editb{0.9912} & \editb{0.8190}\\
BIT & 11.94 & 0.9164 & 0.9170 & \underline{0.9169} & \editb{\underline{0.9930}} & \editb{\underline{0.8466}} & 0.8959 & \underline{0.8965} & 0.8962 & \editb{0.9908} & \editb{0.8119}\\
L-UNet & 8.45 & \underline{0.9318} & 0.8864 & 0.9085 & \editb{0.9919} & \editb{0.8324} & 0.7900 & 0.8938 & 0.8387 & \editb{0.9909} & \editb{0.8144}\\
DSIFN & 50.71 & 0.8906 & \textbf{0.9233} & 0.9067 & \editb{0.9920} & \editb{0.8293} & \textbf{0.9289} & 0.8466 & 0.8859 & \editb{0.9903} & \editb{0.7951}\\
SNUNet & 12.03 & 0.9084 & 0.9110 & 0.9097 & \editb{0.9924} & \editb{0.8343} & 0.8826 & 0.8820 & 0.8823 & \editb{0.9896} & \editb{0.7894}\\
RDP-Net & 1.70 & 0.9148 & 0.8884 & 0.9012 & \editb{0.9913} & \editb{0.8125} & 0.9141 & 0.8893 & \underline{0.9015} & \editb{\underline{0.9913}} & \editb{0.8207}\\
Changer & 11.39 & 0.9249 & 0.9008 & 0.9127 & \editb{0.9923} & \editb{0.8394} & 0.9102 & 0.8855 & 0.8977 & \editb{0.9909} & \editb{0.8144}\\
SiamixFormer & 10.04 & - & - & 0.8947 & \editb{-} & \editb{-} & - & - & 0.8712 & \editb{-} & \editb{-}\\
BAN & 4.34 & 0.9290 & 0.8965 & 0.9124 & \editb{0.9923} & \editb{0.8389} & 0.9180 & 0.8860 & 0.9017 & \editb{0.9913} & \editb{\underline{0.8210}}\\
LightCDNet & 2.82 & 0.9156 & 0.9000 & 0.9077 & \editb{0.9918} & \editb{0.8311} & 0.9083 & 0.8930 & 0.9006 & \editb{0.9912} & \editb{0.8191}\\
\textbf{SRC-Net} & 5.17 & 0.9263 & \underline{0.9172} & \textbf{0.9224} & \editb{\textbf{0.9935}} & \editb{\textbf{0.8560}} & \underline{0.9257} & \textbf{0.9155} & \textbf{0.9206} & \editb{\textbf{0.9930}} & \editb{\textbf{0.8528}}\\
\hline
\hline
\end{tabular}
\end{table*}

Table \ref{tab:result} reports the comparisons of detection accuracy and the number of parameters.
Our proposed SRC-Net shows better performance than other SOTA CD methods with only 5.17M parameters.
On the LEVIR-CD dataset, our SRC-Net achieved the highest F1 of 0.9224, surpassing other SOTA methods.
The precision is 0.9263 and the recall is outstanding 0.9172, both of which are competitive, demonstrating a competitive performance.
BIT secures the second position with an F1 of 0.9169, and 11.94 M parameters.
SRC-Net has about 44\% of BIT's parameters ($\rm 11.94M\times43\%=5.17M$).
On the WHU Building dataset, our SRC-Net also achieved the highest F1 of 0.9224 and the highest recall of 0.9155, surpassing other SOTA methods.
The precision is outstanding at 0.9257, further demonstrating a competitive performance.
RDP-Net secures the second position, boasting an F1 of 0.9169.
Fig. \ref{SOTA} shows some detection results from the validation set of the LEVIR-CD dataset and WHU Building dataset.
\edita{The qualitative results provide a more intuitive comparison of the CD methods.
We observe that most of the comparative methods perform poorly in complex scenes, such as the case of tree occlusion.
In contrast, thanks to the more accurate features extracted by the perception and interaction module and the more expressive features fused by the Patch-Mode joint feature fusion module, our SRC-Net has better performance.}

\begin{figure*}[ht]
\centering
\subfloat[]{
    \includegraphics[width= 0.1\textwidth]{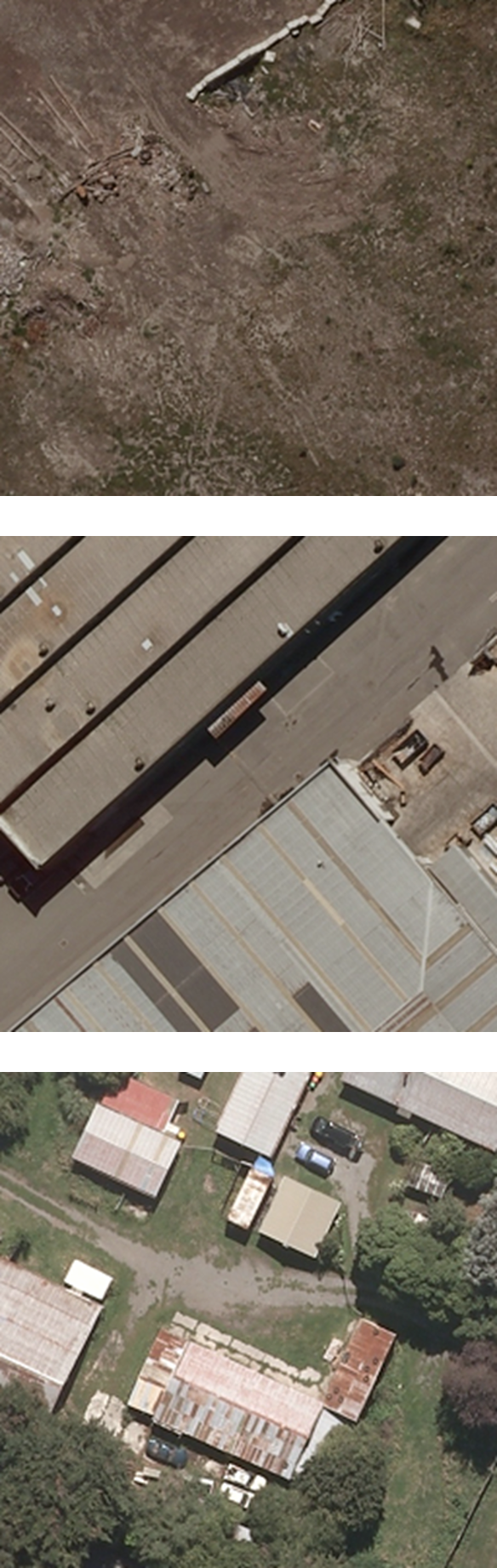}
}
\subfloat[]{
    \includegraphics[width= 0.1\textwidth]{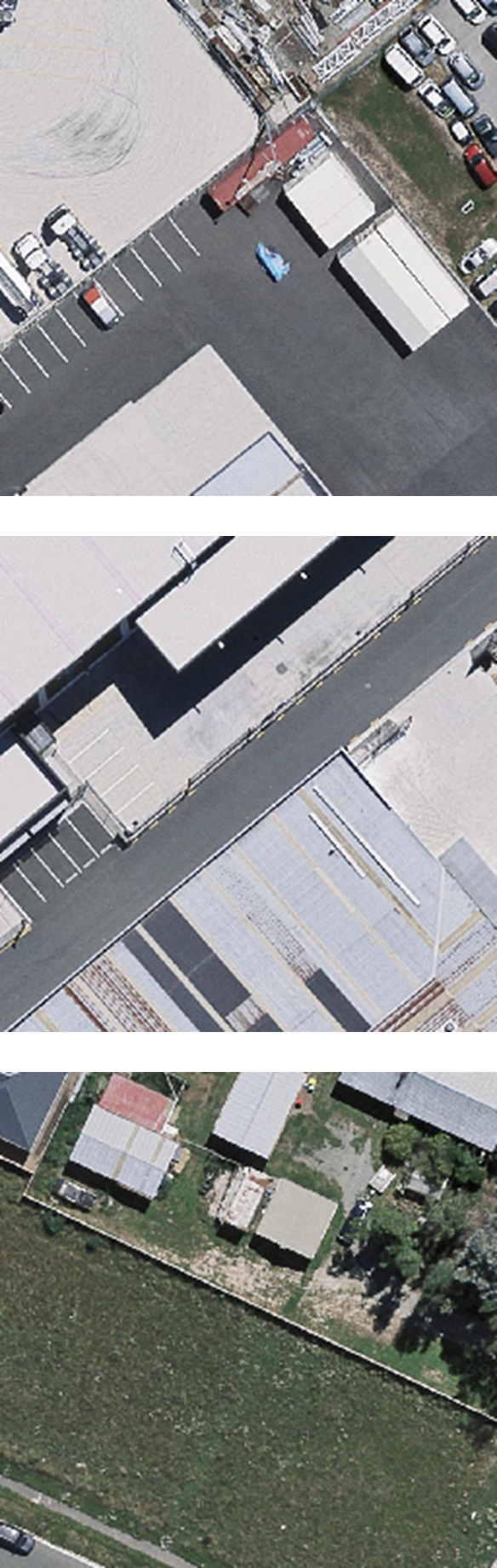}
}
\subfloat[]{
    \includegraphics[width= 0.1\textwidth]{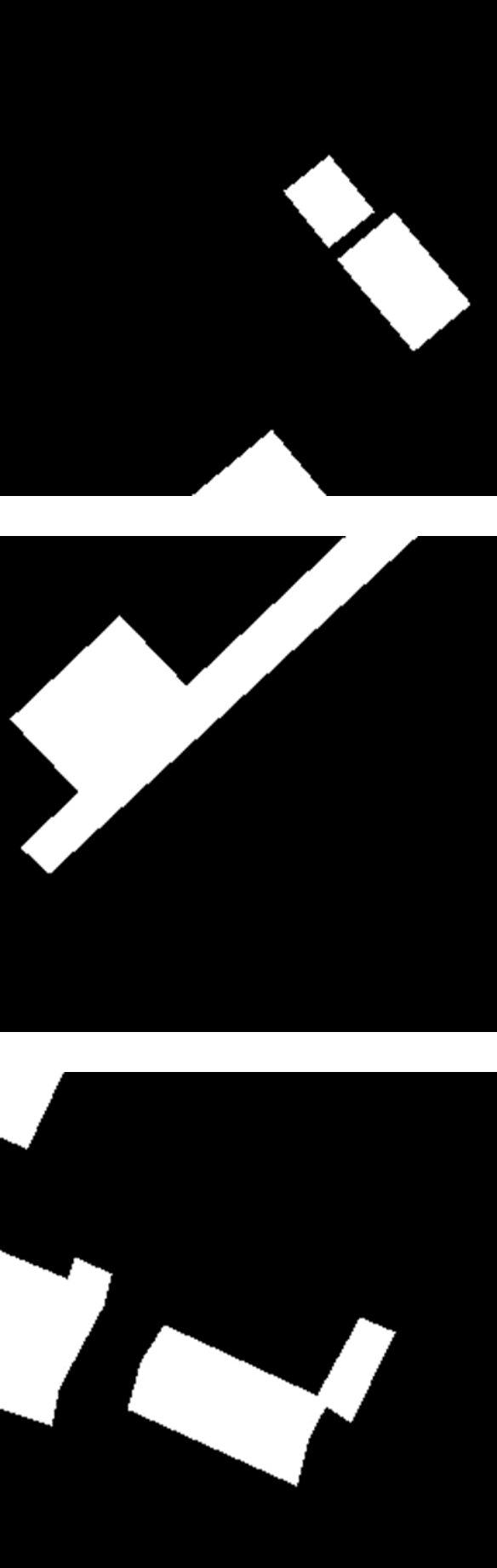}
}
\subfloat[]{
    \includegraphics[width= 0.1\textwidth]{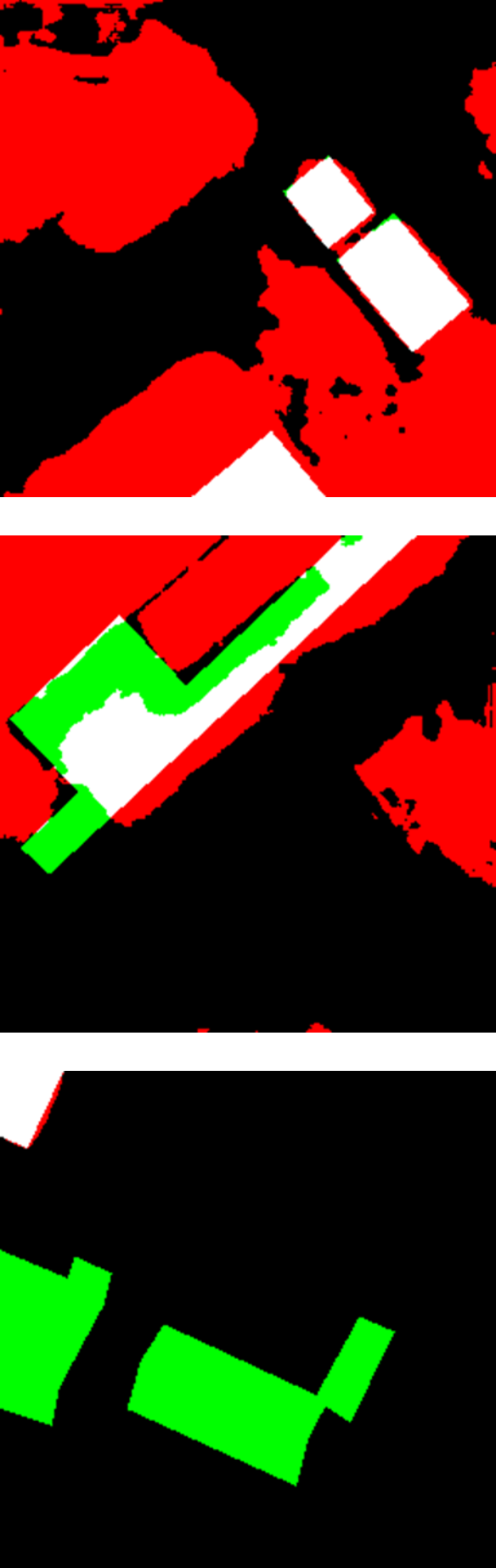}
}
\subfloat[]{
    \includegraphics[width= 0.1\textwidth]{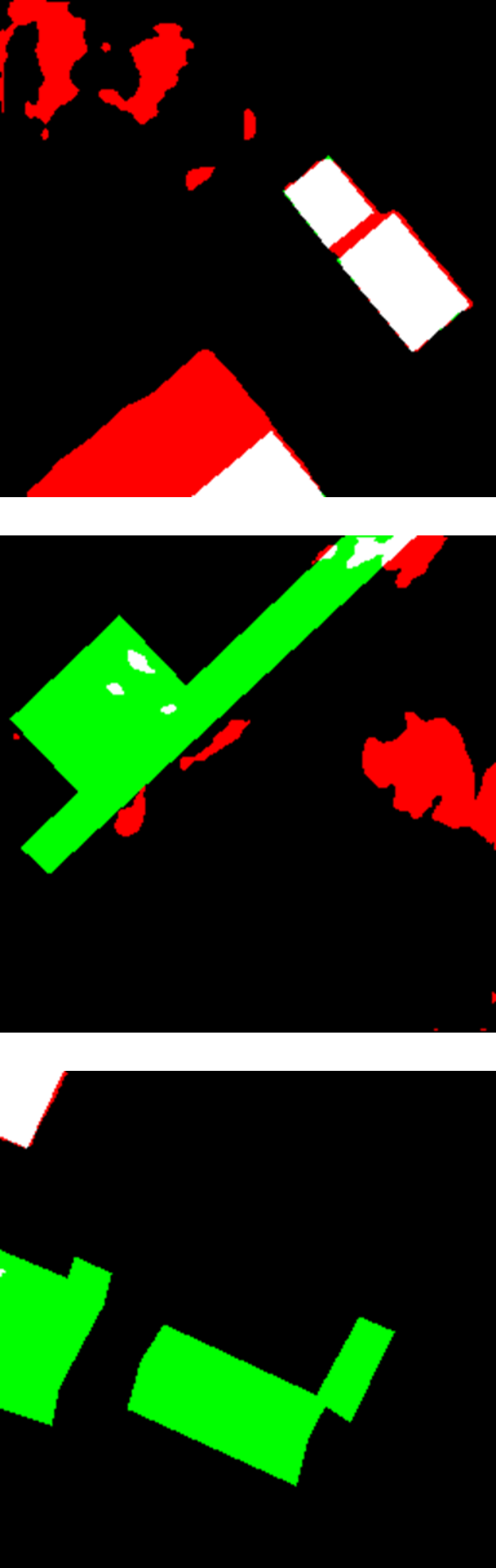}
}
\subfloat[]{
    \includegraphics[width= 0.1\textwidth]{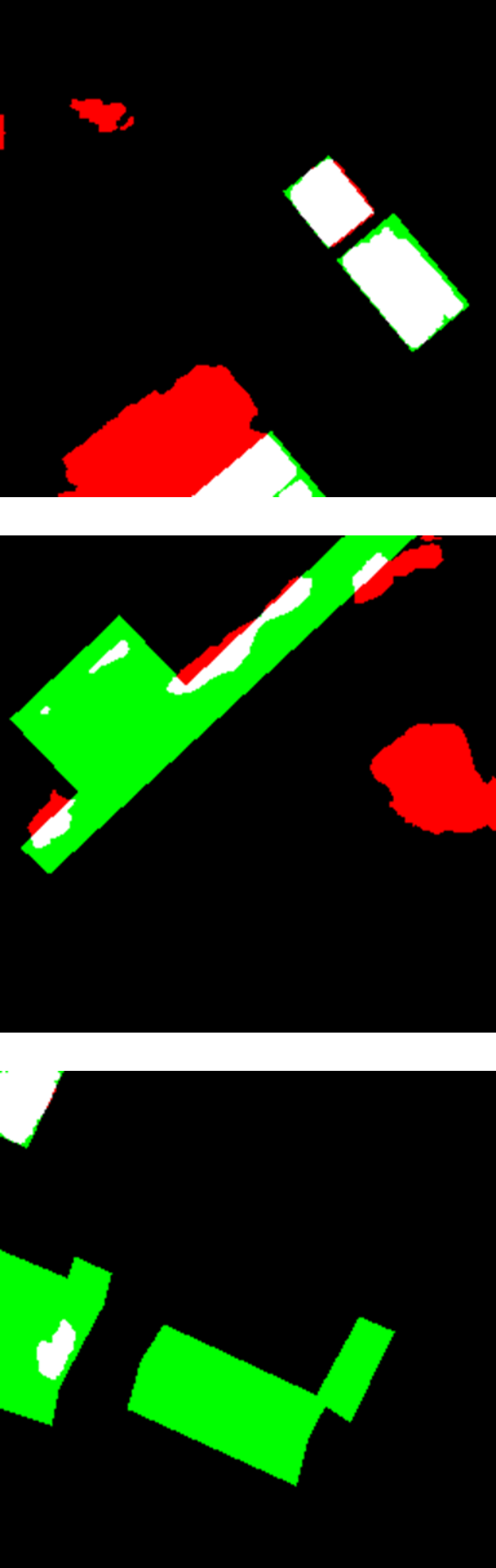}
}
\subfloat[]{
    \includegraphics[width= 0.1\textwidth]{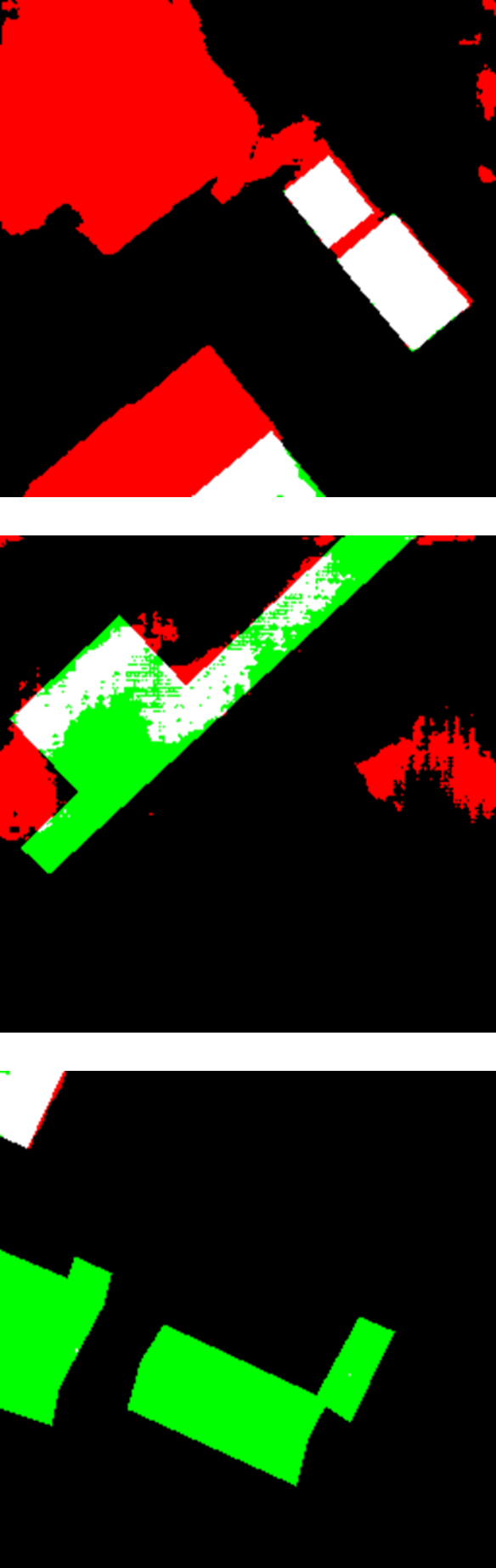}
}
\subfloat[]{
    \includegraphics[width= 0.1\textwidth]{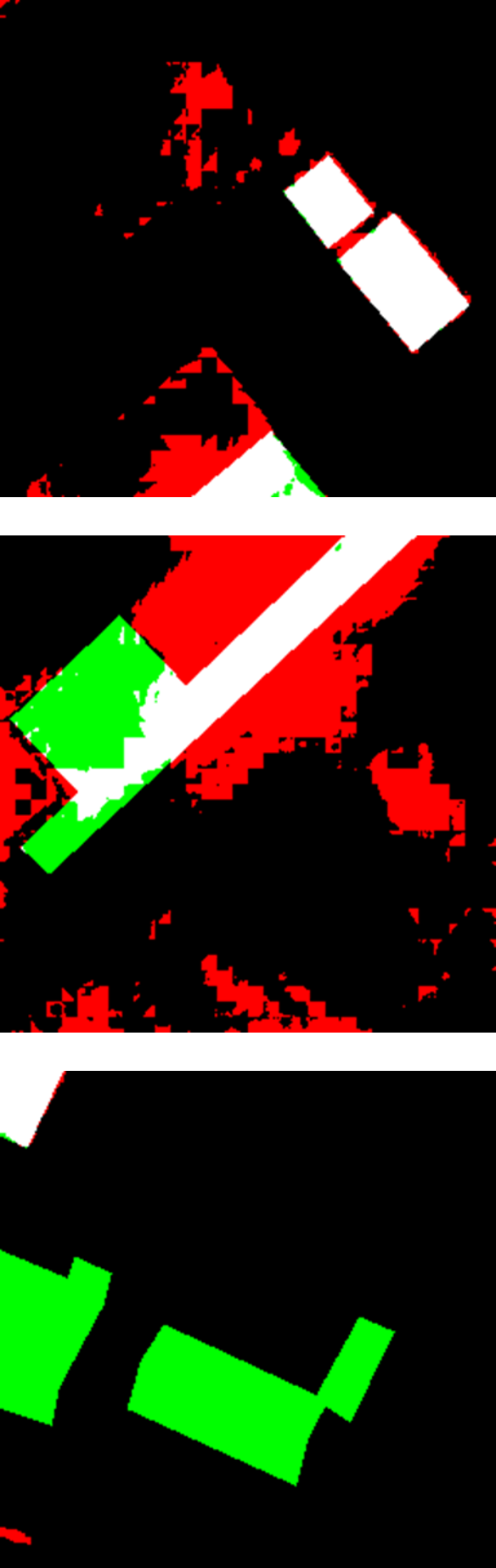}
}
\subfloat[]{
    \includegraphics[width= 0.1\textwidth]{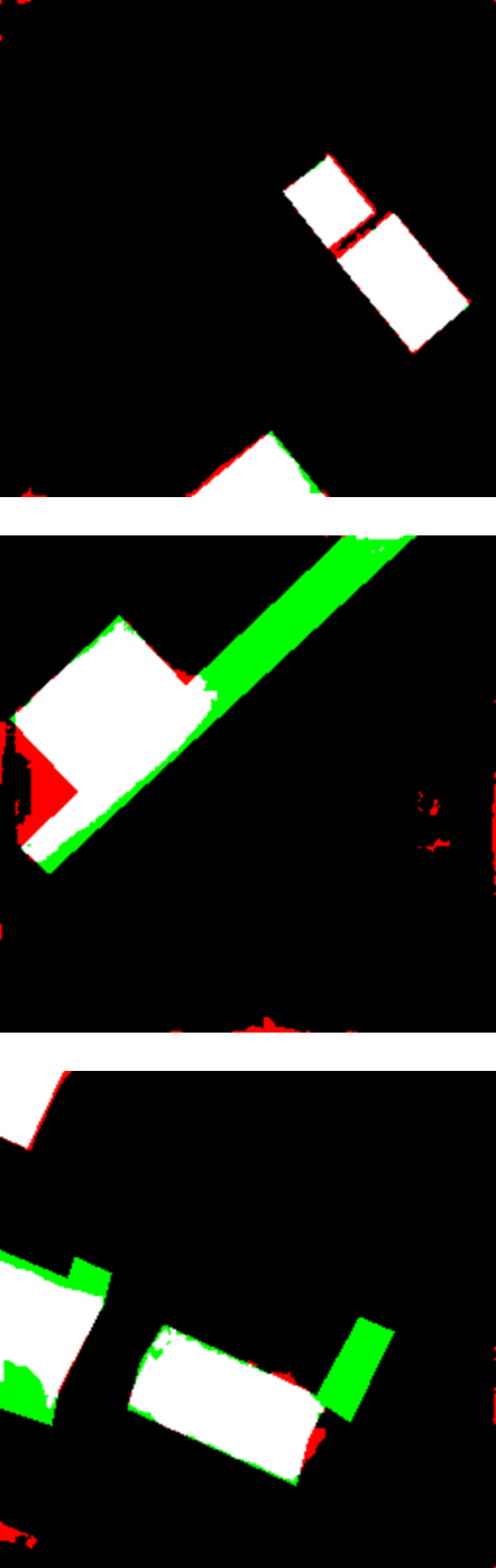}
}
\caption{(a) and (b) are the original images.
(c) are the ground truth.
The result of (d) FC-Siam-diff,
(e) BIT,
(f) DSIFN,
(g) SNUNet,
(h) RDP-Net,
(i) our SRC-Net.
The false positives and false negatives are indicated in red and green, respectively.
Other colors represent true positives.}
\label{update}
\end{figure*}

In our experiments, particularly in certain scenarios, we have also observed significant improvements in our SRC-Net compared to other SOTA methods, as shown in Fig. \ref{update}.
\edita{For scenarios where the optical properties of the road and the roofs of the surrounding buildings are similar, comparative methods perform poorly, and our SRC-Net still gives good results.
However, we also note that there is still room for improvement of our method in this scenario.}
These experiments further prove the superiority of our SRC-Net.
We believe that these improvements are the result of our enhancements, which will be analyzed in detail later.

\subsection{Ablation Study}

To evaluate the perception and interaction module and Patch-Mode joint feature fusion module, a couple of ablation experiments were conducted, as shown in Table \ref{Ablation}.
There are two variants of our network, in addition to the full network. They are:
\begin{enumerate}
\item{\editb{SRC-Net$_\gamma$: We remove perception and interaction module (PIM) from SRC-Net.}}
\item{SRC-Net$_\beta$: We remove the Patch-Mode joint feature fusion module (PM-FFM) from SRC-Net, and just subtract bi-temporal feature maps to obtain the new feature map.}
\item{SRC-Net$_\alpha$: We remove perception and interaction module (PIM) from SRC-Net$_\beta$.}
\end{enumerate}

\begin{table}[ht]
\caption{Ablation Experiment on LEVIR-CD Dataset}
\label{Ablation}
\centering
\begin{tabular}{cccccc}
\hline
\hline
Method & PIM & PM-FFM & Precision & Recall & F1\\
\hline
SRC-Net$_\alpha$ & & & 0.9305 & 0.9025 & 0.9163\\
SRC-Net$_\beta$ & \ding{52} & & 0.9249 & 0.9165 & 0.9206\\
\editb{SRC-Net$_\gamma$} & & \editb{\ding{52}} & \editb{0.9265} & \editb{0.9118} & \editb{0.9191}\\
SRC-Net & \ding{52} & \ding{52} & 0.9263 & 0.9172 & 0.9224\\
\hline
\hline
\end{tabular}
\end{table}

\begin{figure*}[ht]
\centering
\subfloat[]{
    \includegraphics[width= 0.15\textwidth]{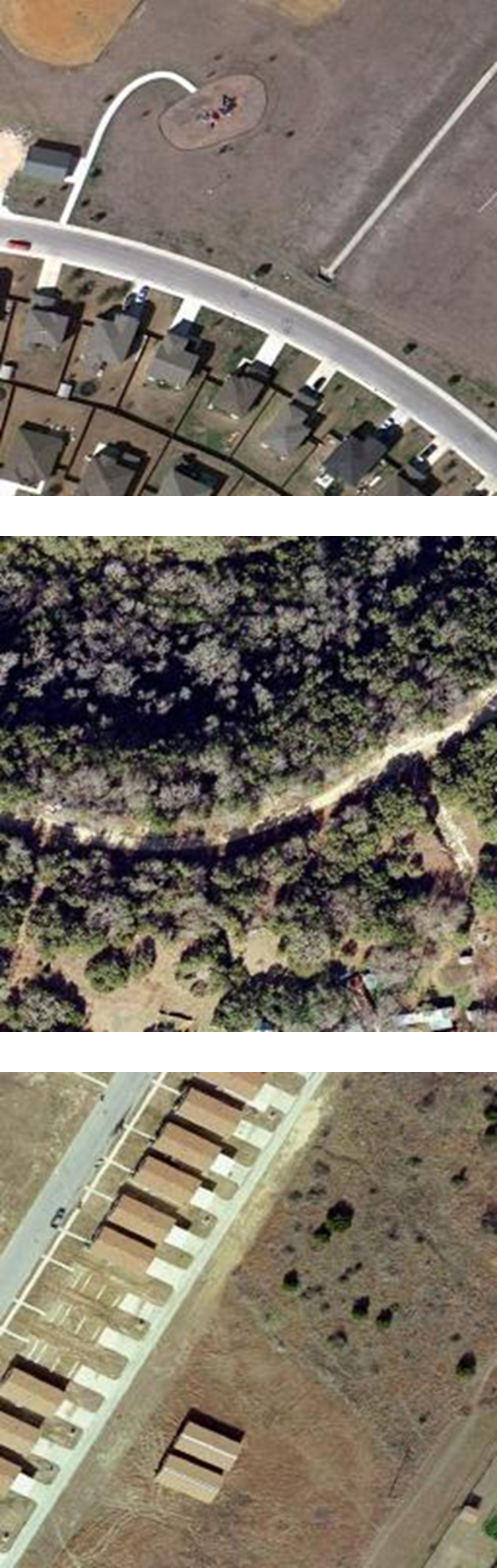}
}
\subfloat[]{
    \includegraphics[width= 0.15\textwidth]{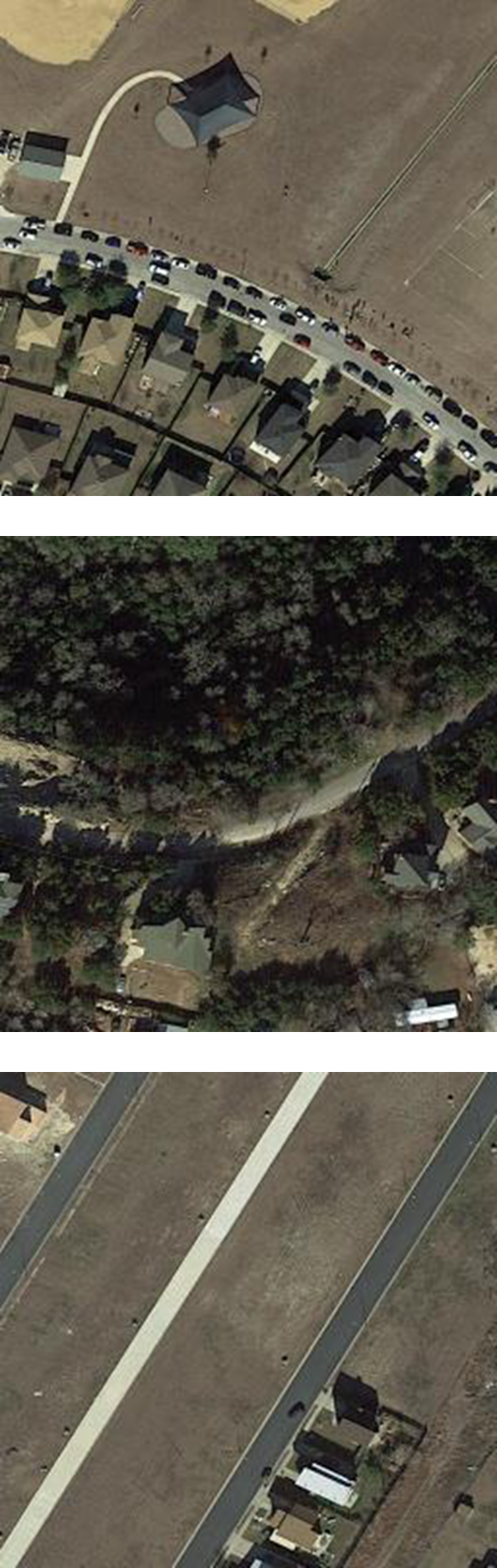}
}
\subfloat[]{
    \includegraphics[width= 0.15\textwidth]{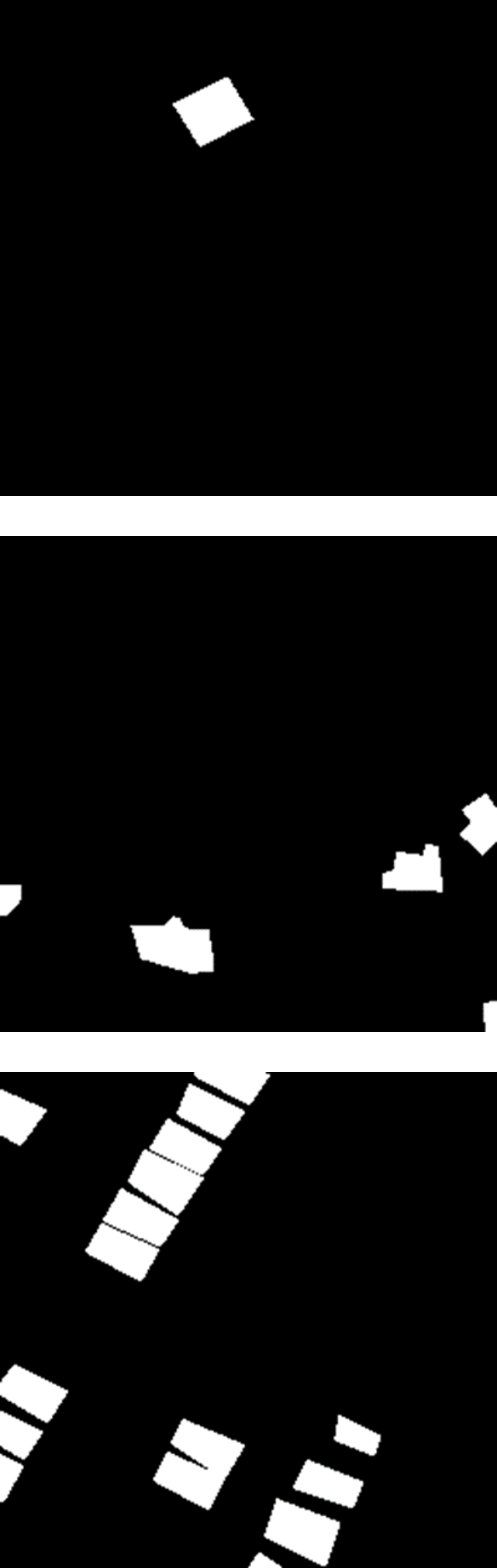}
}
\subfloat[]{
    \includegraphics[width= 0.15\textwidth]{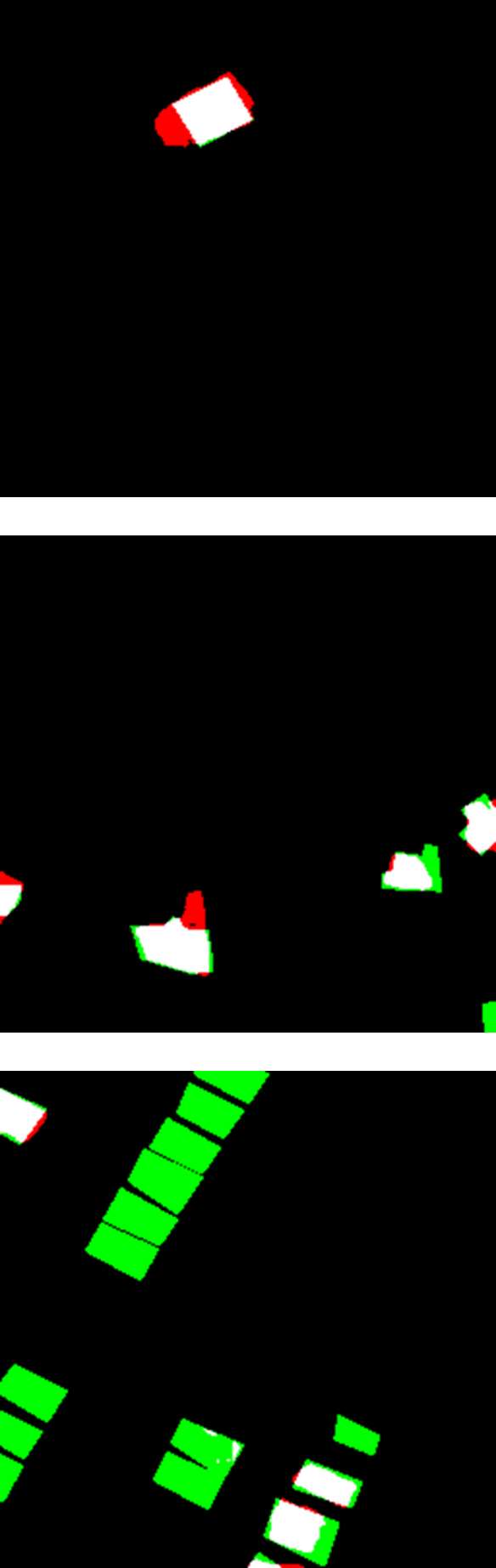}
}
\subfloat[]{
    \includegraphics[width= 0.15\textwidth]{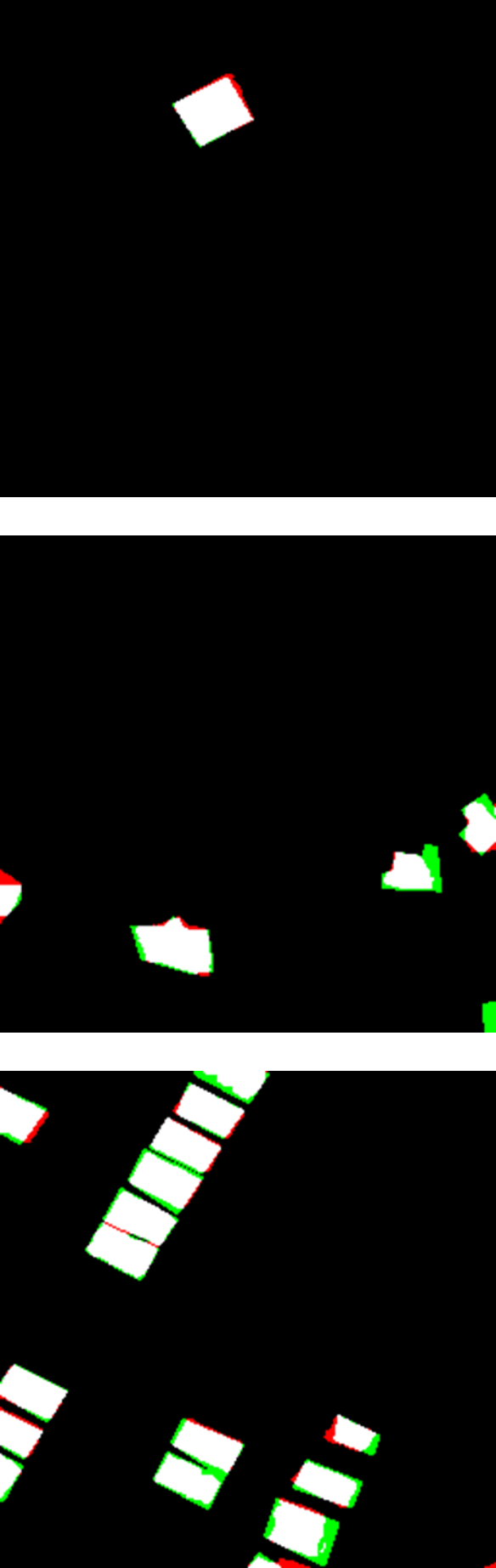}
}
\subfloat[]{
    \includegraphics[width= 0.15\textwidth]{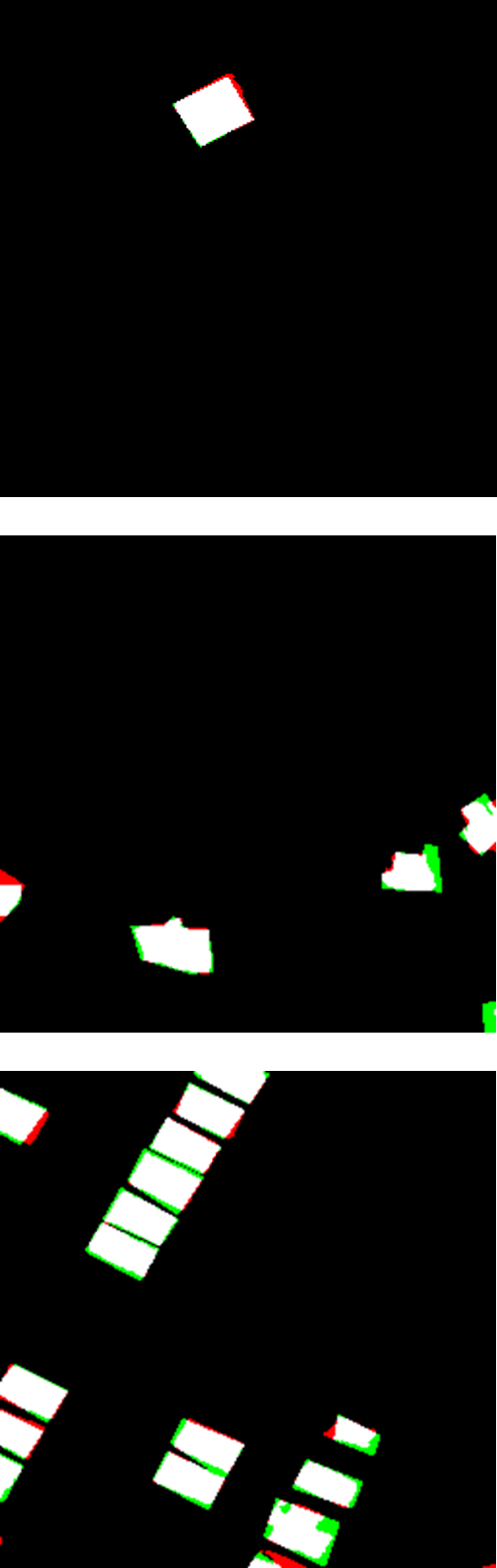}
}
\caption{(a) and (b) are the original images.
(c) are the ground truth.
The result of (d) SRC-Net$_\alpha$,
(e) SRC-Net$_\beta$,
(f) SRC-Net.
The false positives and false negatives are indicated in red and green, respectively.
Other colors represent true positives.}
\label{PIM}
\end{figure*}

The experimental results exemplify the contributions of each improvement in our SRC-Net.
The perception and interaction module (PIM) improves our network by 1.40\% recall and 0.43\% F1.
The Patch-Mode joint feature fusion module (PM-FFM) improves our network by 0.93\% recall, and 0.28\% F1.
Fig. \ref{PIM} shows some detection results.

The experiment results suggest that our improvements can increase recall, thereby allowing the network to detect more change areas.
Additionally, the experiments demonstrate the efficiency of the SRC-Net architecture, enabling the attainment of performance levels comparable to state-of-the-art (SOTA) standards even in the absence of the perception and interaction module and Patch-Mode joint feature fusion module.

\subsection{Experiment in Perception and Interaction Module}

The aforementioned experiments collectively demonstrate the remarkable superiority of our SRC-Net.
By analyzing experimental results, we conclude that a significant factor lies in the better feature extraction capabilities during the bi-temporal geospatial feature extraction stage.
The perception and interaction module facilitates dual-branch communication.
These factors contribute to the network's extraction of more precise and robust features.

\begin{figure}[ht]
\centering
\subfloat[]{
    \includegraphics[width= 0.145\linewidth]{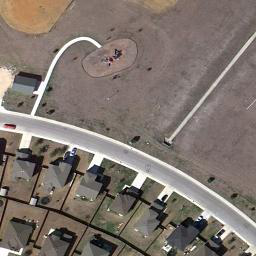}
}
\subfloat[]{
    \includegraphics[width= 0.145\linewidth]{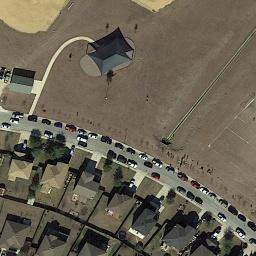}
}
\subfloat[]{
    \includegraphics[width= 0.145\linewidth]{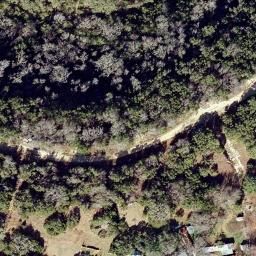}
}
\subfloat[]{
    \includegraphics[width= 0.145\linewidth]{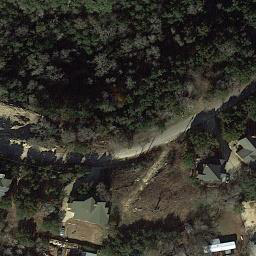}
}
\subfloat[]{
    \includegraphics[width= 0.145\linewidth]{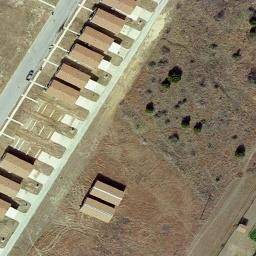}
}
\subfloat[]{
    \includegraphics[width= 0.145\linewidth]{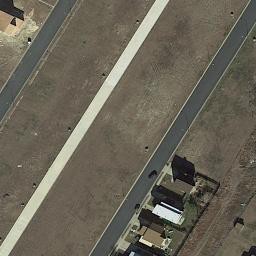}
}

\subfloat[]{
    \includegraphics[width= 0.145\linewidth]{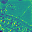}
}
\subfloat[]{
    \includegraphics[width= 0.145\linewidth]{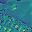}
}
\subfloat[]{
    \includegraphics[width= 0.145\linewidth]{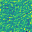}
}
\subfloat[]{
    \includegraphics[width= 0.145\linewidth]{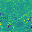}
}
\subfloat[]{
    \includegraphics[width= 0.145\linewidth]{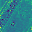}
}
\subfloat[]{
    \includegraphics[width= 0.145\linewidth]{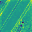}
}

\subfloat[]{
    \includegraphics[width= 0.145\linewidth]{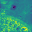}
}
\subfloat[]{
    \includegraphics[width= 0.145\linewidth]{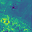}
}
\subfloat[]{
    \includegraphics[width= 0.145\linewidth]{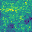}
}
\subfloat[]{
    \includegraphics[width= 0.145\linewidth]{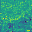}
}
\subfloat[]{
    \includegraphics[width= 0.145\linewidth]{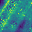}
}
\subfloat[]{
    \includegraphics[width= 0.145\linewidth]{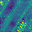}
}
\caption{(a)-(f) are the original images. The visualization results of (g)-(l) the output feature maps of the bi-temporal geospatial feature extraction stage of SRC-Net$_\alpha$, (m)-(r) SRC-Net$_\beta$.}
\label{visual}
\end{figure}

As shown in Fig. \ref{update} and \ref{PIM}, SRC-Net exhibits enhanced discernment when it comes to buildings that are easily confused with their surrounding environments.
We have visualized the bi-temporal feature maps obtained by extracting the instances in Fig. \ref{PIM} through SRC-Net$_\alpha$, SRC-Net$_\beta$, the results are shown in Fig. \ref{visual}.
From the results, it is evident that for regions with changes in buildings, the feature maps extracted by SRC-Net$_\beta$ exhibit more pronounced variations compared to SRC-Net$_\alpha$.
Furthermore, the delineation of the area is more precise.
These are highly advantageous for subsequent predictions of changed areas.

\subsection{Analysis of Patch-Mode joint Feature Fusion Module}

The fusion module is designed to prevent the information loss present in current methods.
Meanwhile, it considers different change modes and concerns about bi-temporal spatial relationships, thereby obtaining more expressive fusion features.

\edita{For the setting of the hyperparameter $k$ in the module, six values from 2 to 64 are selected and tested on the LEVIR-CD dataset.
The results are shown in Table \ref{PMAblation}.
It can be seen that when $k$ is larger than 4 the impact on the final performance will not be significant.
And we set $k$ to 16 based on the result.}

\begin{table}[ht]
\caption{\edita{Experiment with different $k$ in Patch-Mode joint Feature Fusion Module}}
\label{PMAblation}
\centering
\begin{tabular}{ccccccc}
\hline
\hline
\edita{$k$} & \edita{2} & \edita{4} & \edita{8} & \edita{\textbf{16}} & \edita{32} & \edita{64}\\
\hline
\edita{Precision} & \edita{0.9226} & \edita{0.9257} & \edita{0.9236} & \edita{0.9263} & \edita{0.9272} & \edita{0.9267}\\
\edita{Recall}    & \edita{0.9195} & \edita{0.9185} & \edita{0.9209} & \edita{0.9172} & \edita{0.9156} & \edita{0.9164}\\
\edita{F1}        & \edita{0.9210} & \edita{0.9221} & \edita{0.9222} & \edita{0.9224} & \edita{0.9214} & \edita{0.9215}\\
\hline
\hline
\end{tabular}
\end{table}

\begin{figure*}[ht]
\centering
\subfloat[]{
    \includegraphics[width= 0.13\textwidth]{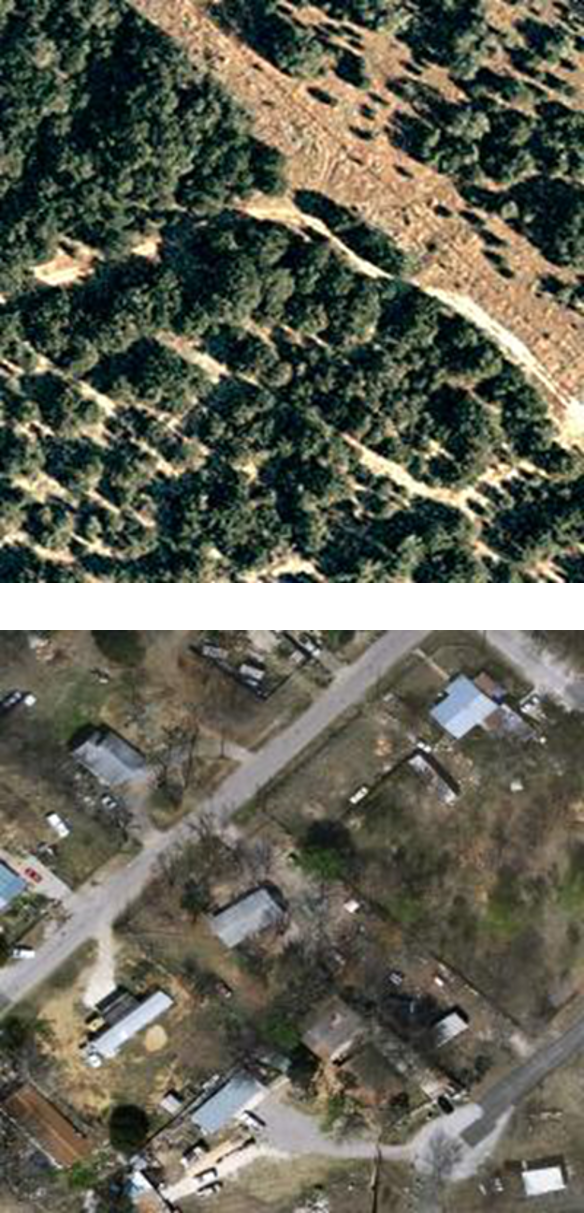}
}
\subfloat[]{
    \includegraphics[width= 0.13\textwidth]{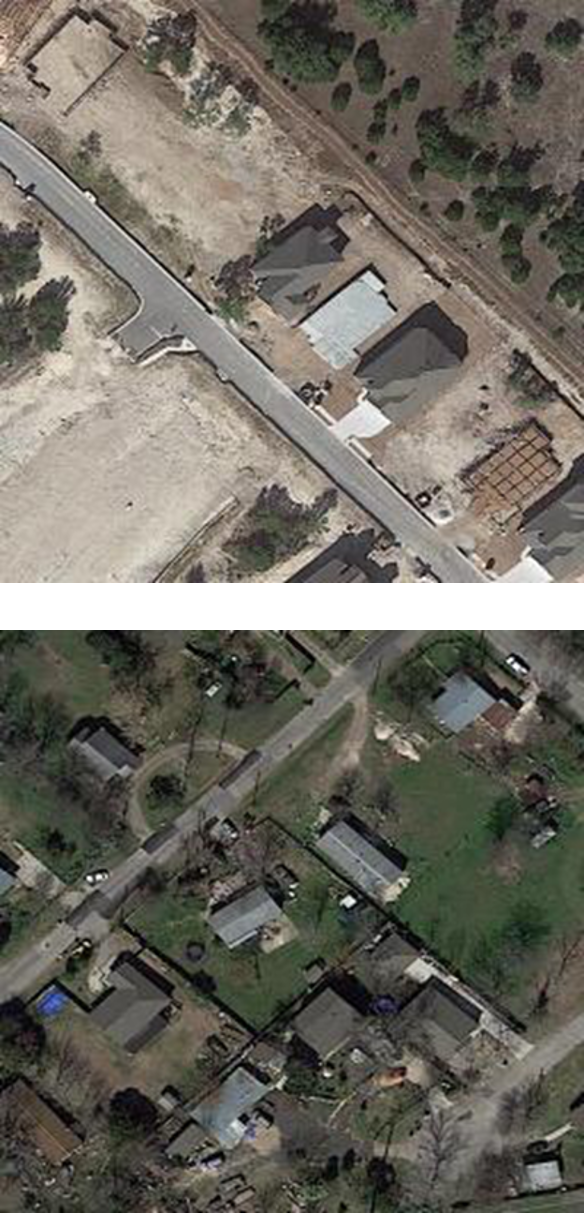}
}
\subfloat[]{
    \includegraphics[width= 0.13\textwidth]{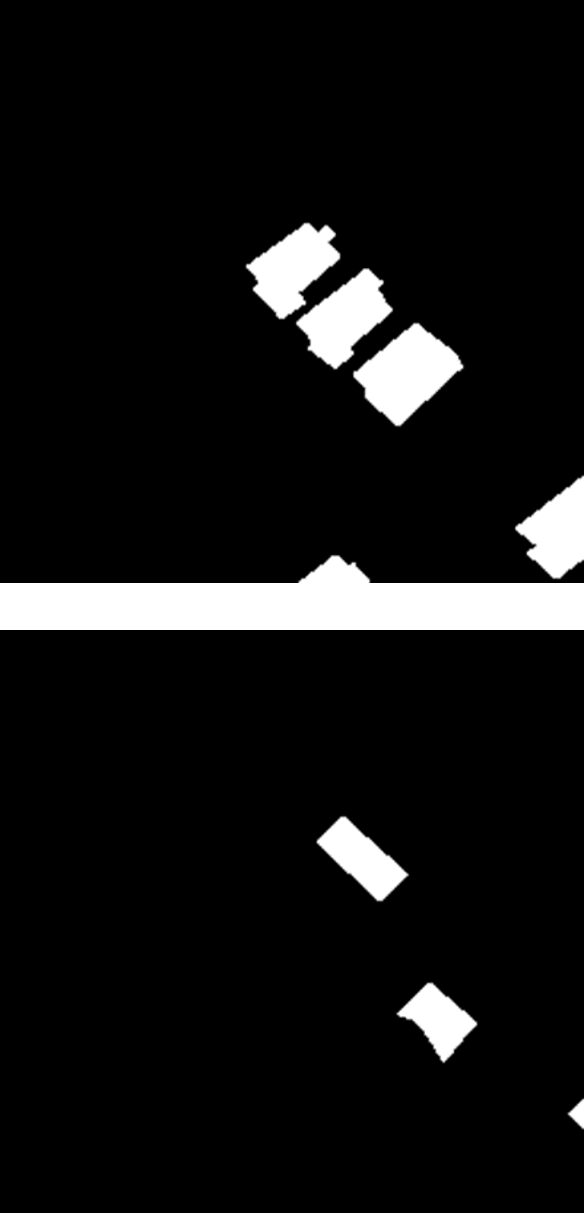}
}
\subfloat[]{
    \includegraphics[width= 0.13\textwidth]{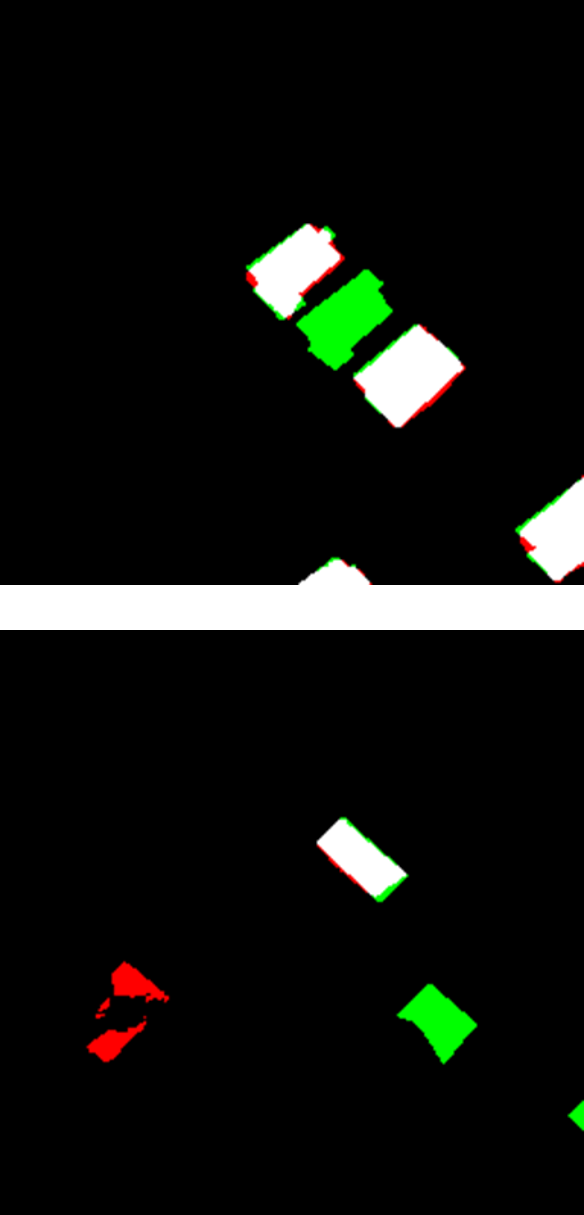}
}
\subfloat[]{
    \includegraphics[width= 0.13\textwidth]{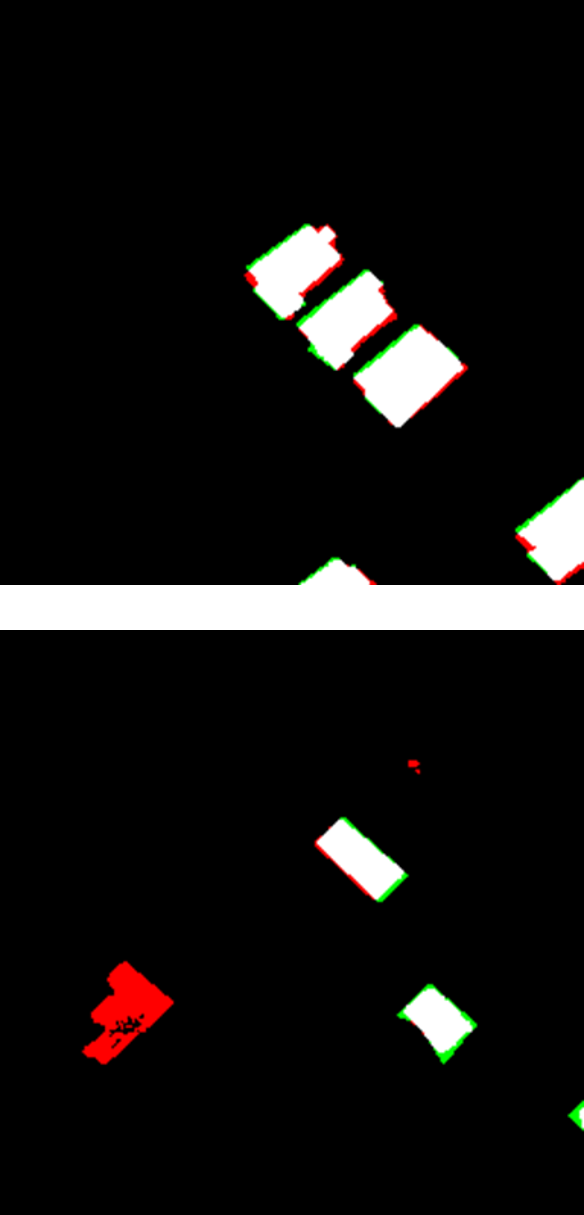}
}
\caption{(a) and (b) are the original images.
(c) are the ground truth.
The result of (d) SRC-Net$_\beta$,
(e) our SRC-Net.
The false positives and false negatives are indicated in red and green, respectively.
Other colors represent true positives.}
\label{ffm}
\end{figure*}

In addition, we visualize the results of bi-temporal feature extraction.
As shown in Fig. \ref{ffm}, in the top row of images, the area surrounding the three houses in the middle remains as land before and after the change.
However, the middle house differs from the other buildings, yet it might be misidentified as flat land when looking at a single image.
If we observe the fusion features by direct subtraction, crucial features that can determine the type of this area may be lost.
As a result, SRC-Net$_\beta$ is unable to identify this particular change. With the RS-CD feature fusion module, this change is well predicted.
In the bottom row, this situation becomes even more apparent.
The input images have significant forest cover, and if we simply subtract them, it becomes challenging for the subsequent network to perceive the surrounding environment's characteristics.
It may fail to recognize the potential noise that could have been introduced in the previous network.

\section{Discussion}
\editb{Most existing CD networks lack structure design for perceiving bi-temporal spatial relationships.
Therefore, existing methods are insufficient in the utilization of bi-temporal spatial relationships, which is crucial for CD.
The SOTA methods compared in Table \ref{tab:result} can be broadly categorized into three groups: single-branch networks (FC-EF, L-UNet, SNUNet, RDP-Net), dual-branch networks without cross-branch interaction (FC-Siam-conc, FC-Siam-diff, STANet, SiamixFormer, BAN), and dual-branch networks with cross-branch interaction (BIT, DSIFN, Changer, LightCDNet, SRC-Net).
From the results, it can be seen that dual-branch networks without cross-branch interaction achieve generally poorer results than single-branch networks, while dual-branch networks with cross-branch interaction achieve better results.
The results further validate the importance of bi-temporal spatial relationship.
The two modules proposed in this paper significantly enhance the network's ability to perceive spatial correlation during the feature extraction and feature fusion stages.}

\editb{However, the current SRC-Net still has some limitations.
The SRC-Block can be reduced in the amount of computation and the number of parameters.
Additionally, the perception and interaction module relies on the learned credibility matrices for interaction, ignoring that the land cover may have changed between the two time phases.
These issues will be the focus of our future research.}

\section{Conclusion}
In this article, we propose SRC-Net, an effective method concerning bi-temporal spatial relationships for remote sensing CD.
It contains the perception and interaction module and the Patch-Mode joint feature fusion module.
These modules effectively utilize the bi-temporal spatial relationships in the input bi-temporal data.
The perception and interaction module establishes a cross-branch perception mechanism during the feature extraction process with the bi-temporal spatial relationships among bi-temporal inputs, leading to extract more precise and robust features.
The Patch-Mode joint feature fusion module considers different change modes, and concerns bi-temporal spatial relationships, and prevents information loss in current methods.
With this module, we can obtain more expressive fusion features.
We have corroborated the effectiveness of our network through comprehensive experiments.
The proposed SRC-Net achieves the SOTA empirical performance with 5.17M parameters.
Experiments were conducted on the LEVIR-CD and WHU Building datasets, with the results indicating that the SRC-Net model has the potential to notably improve the accuracy of remote sensing CD.
\edita{However, we also note that the computational overhead of the current backbone of SRC-Net remains a limiting factor for deployment on edge devices.
The research in this paper is based on bi-temporal remote sensing images, but the two modules proposed can be also applied in multi-temporal CD tasks, which will be the direction of our subsequent research.}

\bibliography{reference}
\bibliographystyle{IEEEtran}

\vfill

\end{document}